%% file: paper.tex
\documentclass[10pt]{article}
\usepackage[preprint]{tmlr}
\input{math_commands.tex}

\usepackage[utf8]{inputenc}
\usepackage[T1]{fontenc}
\usepackage{lmodern}
\usepackage{hyperref}
\usepackage{url}
\usepackage{graphicx}
\usepackage{placeins}
\usepackage{booktabs,makecell,tabularx}
\usepackage{amssymb}
\providecommand{\tightlist}{\setlength{\itemsep}{0pt}\setlength{\parskip}{0pt}}
\DeclareUnicodeCharacter{2212}{\ensuremath{-}}
\DeclareUnicodeCharacter{2248}{\ensuremath{\approx}}
\DeclareUnicodeCharacter{00D7}{\ensuremath{\times}}
\DeclareUnicodeCharacter{03BB}{\ensuremath{\lambda}}
\DeclareUnicodeCharacter{03BA}{\ensuremath{\kappa}}
\DeclareUnicodeCharacter{0394}{\ensuremath{\Delta}}
\DeclareUnicodeCharacter{2265}{\ensuremath{\geq}}
\DeclareUnicodeCharacter{2264}{\ensuremath{\leq}}
\DeclareUnicodeCharacter{2192}{\ensuremath{\rightarrow}}
\DeclareUnicodeCharacter{03C4}{\ensuremath{\tau}}
\DeclareUnicodeCharacter{00B1}{\ensuremath{\pm}}
\DeclareUnicodeCharacter{2009}{\,}
\DeclareUnicodeCharacter{00A0}{~}
\DeclareUnicodeCharacter{03B2}{\ensuremath{\beta}}
\DeclareUnicodeCharacter{03C0}{\ensuremath{\pi}}
\DeclareUnicodeCharacter{0233}{\ensuremath{\bar{y}}}
\DeclareUnicodeCharacter{207A}{\ensuremath{^{+}}}
\DeclareUnicodeCharacter{2080}{\ensuremath{_{0}}}
\DeclareUnicodeCharacter{2081}{\ensuremath{_{1}}}
\title{Abstention Is Not Threshold Compliance: Stated Scoring Rules in Open Language Models}
\author{\name Dongsun Moon \email dongsunmoon@gmail.com \\ \addr Independent Researcher}
\begin{document}
\maketitle
\suppressfloats[t]
\begin{abstract}
Kalai et al.~propose stating the scoring rule in the prompt with an explicit threshold \(t\), so that a model should answer only when its probability of being correct exceeds \(t\). We probe each requirement of compliance on shared items, in 12 checkpoints from three post-training stages of four fully open 3--8B recipes, with exploratory models up to 32B. At \(t\) = 0.75, a line limiting the reply to an answer or ``I abstain'' (the abstention format) raised the mean abstention rate from 20\% to 91\%. Yet within this format, raising \(t\) from 0.5 to 0.9 changed it by only −0.8 percentage points. Without the line, the higher threshold even lowered abstention, and up to 40\% of a checkpoint's items received decisions that no compliant policy with threshold-independent beliefs could make. Given the probability that their answer was correct, the checkpoints barely responded to the threshold in a forced-choice probe. Comparisons with self-assessment signals such as P(True) depended on the grader and on answer drift between prompts. An external threshold on recalibrated P(True) made coverage non-increasing in \(t\) by construction. Calibrated on the evaluation distribution (post hoc), it beat the abstention format at \(t\) = 0.5; calibrated on TriviaQA and applied to PopQA, however, it fell below that format at 0.75. On OLMo 3 7B, decisions trained on threshold-conditioned targets did not follow unseen thresholds consistently. A post-hoc variant trained at four thresholds answered less as \(t\) rose but still failed a pre-specified test, mainly on agreement with its training rule. A high abstention rate is not evidence of compliance: evaluations using stated scoring rules should report how often models abstain, whether decisions respond to the stated threshold, and which items they answer.
\end{abstract}
\section{Introduction}\label{introduction}

Under binary grading, a guess never scores worse than an abstention, so evaluations that reward accuracy alone reward guessing over acknowledging uncertainty \citep{kalai2026evaluating}. Kalai et al.~propose stating the scoring rule in the prompt with an explicit threshold \(t\): an abstention earns \(t\), a correct answer 1 and a wrong answer 0, so a model should answer only when its probability of being correct exceeds \(t\). In their case study, four frontier models abstained more as \(t\) rose, yet all scored below always-abstain at 0.75 and 0.9.

\begin{figure}[t]
\centering
\includegraphics[width=0.82\textwidth]{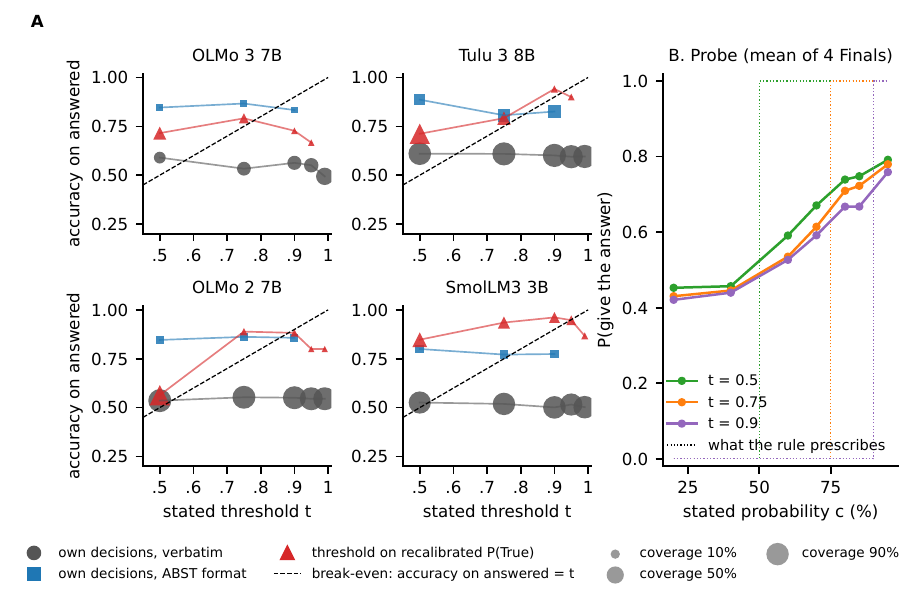}
\caption{\textbf{Decisions under the stated rule against an external threshold.} (A) Accuracy on answered items against the stated threshold \(t\) for the four final checkpoints on the same 500 items: own decisions under the verbatim rule and with the ``I abstain'' line (ABST), and the never-abstain (OR@0) answer given when cross-fitted, recalibrated P(True) exceeds \(t\). Marker area represents coverage; above the break-even line (accuracy = \(t\)) a policy gains over abstaining, by \(R_t - t\) = coverage × (accuracy − \(t\)). (B) Probe: probability of giving the candidate answer against the stated probability \(c\); dotted steps show the rule.}\label{fig1_where_threshold}
\end{figure}

When abstention rises under such a rule, it is tempting to read it as compliance. It need not be. Following a stated threshold requires three things: (i) abstention must be an available reply; (ii) the threshold must be applied to a probability of being correct; and (iii) the model must have a usable estimate of that probability and use it. An instruction that names the abstention reply addresses only (i), yet can raise abstention by itself, so an evaluation reporting only abstention rates cannot tell compliance from a response to the wording.

Prior work has shown parts of this picture. Stated stakes moved abstention less than reward maximization requires, or not at all \citep{wu2025answer,wang2026are}, and thresholds or penalties in the prompt moved it only partly \citep{kumaran2026causal,mohamadi2025honesty}. Instructions and answer options moved it strongly \citep{zong2026icalm,ling2026llm,kim2026penalty}, and thresholds applied afterwards to confidence improved utility \citep{wang2026are,wu2025mitigating}; we claim neither of these as new.

This study introduces a single design that probes each requirement of compliance on the same short-answer items, in fully open checkpoints whose post-training stages are public, so that one evaluation can tell a rise in abstention from compliance. We cross the abstention format, a line limiting the reply to an answer or ``I abstain'', with the stated threshold and check each item's decisions across thresholds (Section 4.2). We then supply the probability of being correct (Section 4.3), compare decisions with self-assessment signals at matched coverage (Section 4.4) and with a threshold applied outside the model (Section 4.5), and compare post-training stages (Section 4.6). Finally, we test interventions that place the threshold in a trained decision, in trained confidence or outside the model (Section 5).

Our contributions are two.

\begin{enumerate}
\def\labelenumi{\arabic{enumi}.}
\tightlist
\item
  \textbf{Abstention and compliance, measured separately.} The abstention format raised mean abstention from 20\% to 91\% at \(t\) = 0.75, while raising \(t\) from 0.5 to 0.9 changed it by −0.8 percentage points. Without that line, the higher threshold even lowered abstention, by 5.5 percentage points, and up to 40\% of a checkpoint's items received decisions that no compliant policy with threshold-independent beliefs could make. Given the probability that their answer was correct, the 3--8B checkpoints barely responded to the threshold in a forced-choice probe (Sections 4.2--4.3; Figure 1B).
\item
  \textbf{Control through an external threshold, with limits.} A threshold on recalibrated P(True) makes coverage non-increasing in \(t\) by construction; with labels from the evaluation items (post hoc), its advantage over the format was clear at \(t\) = 0.5, uncertain at 0.75 and, on the reward scale, small at 0.9. Transferred from TriviaQA to PopQA, it fell below that format at 0.75 (Section 4.5).
\end{enumerate}

Two diagnostic lessons accompany these. Comparisons between decisions and self-assessment signals were fragile: they changed with the handling of answers that differed between prompts, with the checkpoint, and when a deterministic grader replaced the LLM grader (Section 4.4). And on OLMo 3, a decision trained on threshold-conditioned targets did not follow unseen thresholds consistently and lost its abstention when one line was added to the prompt; trained at four thresholds instead of two (post hoc, one seed), it tracked \(t\) in coverage but still failed a pre-specified test, mainly on agreement with its training rule (Section 5.2). Confirmatory analyses were fixed before grading, and post-hoc analyses are labelled as such. Code, prompts, item ids and outputs are in the anonymized supplement; item texts are rebuilt from the original datasets.

\section{Related work}\label{related-work}

\textbf{Stated scoring rules.} \citet{wu2025answer} stated rewards and penalties for multiple-choice items: frontier models refused more as the penalty grew but deviated from reward maximization, and made better choices when the decision was posed alone or split into answering, confidence estimation and expected-value reasoning, close to our requirements. RiskEval \citep{wang2026are}, asking, like Section 4.4, whether decisions follow verbal confidence, found abstention nearly flat across stated penalties, and utility gains from thresholding verbal confidence afterwards. \citet{kumaran2026causal} found large models abstained more as an instructed threshold rose but departed from it: GPT-4o abstained on 47\% of items at a 10\% threshold. I-CALM \citep{zong2026icalm} found selective answering far more sensitive to the reward for abstaining than to the error penalty. In Kalai et al.'s wording \(t\) is that reward, yet raising it barely moved abstention (Section 4.2); the two fit if, as we read I-CALM's Table 10, whether abstaining is rewarded matters more than how much.

\textbf{Format effects.} How abstention is offered can move it more than the stakes do: an explicit ``I don't know'' instruction collapsed coverage for Llama-3-8B \citep{zong2026icalm}, an added ``Unknown'' option inflated abstention on true/false items \citep{ling2026llm}, and an instruction to abstain moved abstention more than the penalty size \citep{kim2026penalty}. AbstentionBench \citep{kirichenko2025abstentionbench} measures abstention on unanswerable questions, including across Tulu 3's stages; we ask, on answerable questions, whether the stated threshold moves decisions once abstention is named.

\textbf{Selective prediction and self-assessment.} Answering only when the probability of being correct exceeds a cost threshold is an optimal reject rule \citep{chow1970on}, and comparing policies at matched coverage follows selective classification \citep{elyaniv2010on,geifman2017selective}. \citet{kadavath2022language} introduced P(True) and P(IK), the latter as a trained value head; our P(IK) is prompted. Verbalized confidence tends to be overconfident \citep{xiong2024can}, hidden states carry extra error information \citep{orgad2025llms}, and confidence predicts and, in one model, causally drives abstention \citep{kumaran2026causal}.

\textbf{Training abstention and external thresholds.} Fine-tuning has taught models to decline questions beyond their knowledge \citep{zhang2024rtuning,cheng2024can,yang2024alignment} and to state calibrated confidence \citep{lin2022teaching,kapoor2024large}. With \(t\) in the prompt, reinforcement learning left decisions insensitive to \(t\) \citep{wu2025mitigating}, and a single model conditioned on the penalty responded weakly, while models trained for one penalty each beat an external threshold at high penalties \citep{mohamadi2025honesty}. Reinforcement learning with abstention as an action can collapse to always abstaining \citep{che2026abstention}, and abstention has also been set from label-free signals or conformal thresholds \citep{asaria2026can,tayebati2025cap}. Our external threshold, the simplest of these, is a reference, not a proposed method.

\textbf{Post-training and calibration.} RLHF policies looked miscalibrated until one temperature was tuned \citep{kadavath2022language}, confident errors grew across Tulu 3's stages \citep{huang2026alignment}, and later training can raise or lower the usefulness of confidence \citep{li2026post,flores2026confident}.

\section{Setup}\label{setup}

\subsection{What compliance with a stated threshold requires}\label{what-compliance-with-a-stated-threshold-requires}

Under the rule of \citet{kalai2026evaluating}, answering pays exactly when the probability of being correct exceeds \(t\) (Section 1; scores in Section 3.5). We call a policy \emph{threshold-compliant} if it answers exactly when its own probability of being correct exceeds \(t\). Because that probability is not observed, we test what compliance implies under threshold-independent beliefs: coverage (the share of items answered) cannot rise with \(t\), and no item declined at one threshold can be answered at a higher one. Passing these checks is necessary for compliance, not sufficient. Compliance that pays off needs three things: (i) abstention is an available reply; (ii) the threshold is applied to a probability of being correct; and (iii) the model has a usable estimate of that probability and uses it. Each maps to a quantity an evaluation can report: (i) how often the model abstains (Section 4.2); (ii) whether its decisions respond to \(t\) consistently (Section 4.2) and whether it compares a supplied probability with \(t\) (Section 4.3); and (iii) which items it answers at matched coverage (Section 4.4).

\subsection{Models}\label{models}

The primary set is 12 checkpoints: the supervised fine-tuning (SFT), preference-optimization (PO) and final (released) stages of OLMo 3 7B Instruct, OLMo 2 7B, Tulu 3 8B and SmolLM3 3B \citep{olmo2025olmo3,olmo2024olmo2,lambert2025tulu,bakouch2025smollm3}. They are related: OLMo 2 and Tulu 3 share nearly the same post-training recipe, and the preference data come, in effect, from two lineages. Exploratory arms add two 32B models run in 8 bits (OLMo 2 32B and OLMo 3.1 32B) and six other open 7--14B models (\citealp{qwen2024qwen25}; \citealp{qwen2025qwen3}; \citealp{jiang2023mistral}; \citealp{abdin2024phi4}; \citealp{granite2025granite}; Appendix Table A6). They also add two reasoning models, OLMo 3 Think and RL-Zero General (Appendix Table A11), and SmolLM3's mid-training checkpoint (Appendix Table A12). All weights are public and pinned to commits; decoding is greedy, with at most 256 new tokens (more for the reasoning models; Appendix B).

\subsection{Items and conditions}\label{items-and-conditions}

The items are 1,000 short-answer questions chosen by a seeded rule: 500 from TriviaQA \citep{joshi2017triviaqa} and 500 from PopQA \citep{mallen2023when}, 50 per popularity decile. Five hundred MMLU-Redux 2.0 items \citep{gema2025are} measure multiple-choice confidence, and the first 200 of 300 sampled SimpleQA Verified items \citep{haas2025simpleqa} place our models beside Kalai et al.'s (Appendix Table A9). Excluding the 86 questions that overlap a recipe's released post-training data changes no hypothesis decision. Prompts follow Kalai et al.'s template (Appendix B). The verbatim rule, OR@\(t\) (open rubric at \(t\)), states the rule at threshold \(t\) (``{[}Abstain if probability of being correct is \textless{} \(100t\)\%. Abstention is scored at \(100t\)\% correct.{]}''), OR@0 tells the model never to abstain, and C states no rule. The abstention format (ABST) adds the line ``Reply with only your answer, or with `I abstain'.'' A pre-freeze pilot had shown its effect (Appendix E). The verbatim rule was stated at \(t\) = 0.5, 0.75, 0.9, 0.95 and 0.99 on all items. On 500 of them, 250 per dataset (the variant items), the verbatim format (without that line) and ABST were crossed with no rule and at \(t\) = 0.5, 0.75 and 0.9; at 0.75 these items also received four wording variants and a permission line without a threshold (P).

\subsection{Self-assessment signals and the probe}\label{self-assessment-signals-and-the-probe}

A model's OR@0 reply is its belief answer, and \(y\) is the primary grader's verdict on it. Stated confidence \(s\) is the probability of being correct that the model reports in a second turn, as a proportion. P(True) asks, in a fresh chat, whether the belief answer is correct, and P(IK) asks, before any answer, whether the model knows the answer; both are read from the two option letters' probabilities, averaged over both orders, and neither is a probability of being correct until recalibrated on labelled items (Section 4.5). The probe targets (ii). On 50 items it shows the model's own belief answer as a candidate, states ``The probability that the candidate answer is correct is \(c\)\%'', adds the OR@\(t\) rule and asks whether to give the answer or abstain. It does so for \(c \in \{20, 40, 60, 70, 80, 85, 95\}\) with no rule and with \(t\) = 0.5, 0.75 and 0.9; \(P_{\text{give}}\) is the probability of the ``give'' letter. An abstract version poses the same grid without a question.

\subsection{Scores and measures}\label{scores-and-measures}

\textbf{Scores.} Each reply is graded correct, wrong or abstention (Section 3.6; notation, and what each score holds fixed, in Appendix A). At threshold \(t\), the utility of \(n\) replies is \(U_t = (n_{\text{correct}} - \tfrac{t}{1-t}\, n_{\text{wrong}})/n\), and always-abstain scores 0. On the abstention-reward scale of the prompt, \(R_t = t + (1-t)U_t\), and \(R_t - t = (1-t)U_t\) is the score above always-abstain. Up to scaling and a correctness-only term, \(U_t\) is the elementary threshold score of which proper scoring rules are mixtures \citep{schervish1989a,ehm2016of}. The potential gain above always-abstain shrinks as \(t\) grows, so we give both scales where a reading depends on it. Scores are \emph{end to end}: each policy is scored on the replies it gives. Two \emph{decision-only} scores hold the answer content fixed instead, scoring answered items by \(y\), the belief answer's grade. Section 4's \(R^{\mathrm{dec}}_t\) codes abstentions as the primary grader does (its mean is within 0.003 of \(R_t\); Appendix Table A3). Section 5's declared-decision score \(D_t\) (fixed in advance) is on the \(U\) scale and codes an abstention when the reply's first sentence contains an abstention phrase (the first-sentence rule). For OLMo 3 Final at \(t\) = 0.75 this coding alone accounts for 0.273 of the 0.309 gap between \(U\) and \(D\) (Appendix Table A7), so we report both.

\textbf{Measures.}

\begin{itemize}
\tightlist
\item
  \emph{Threshold responsiveness} (probe): \(E_{\text{DiD}}\), a difference in differences, is the mean over \(c \in \{60, 70, 80, 85\}\) of \(P_{\text{give}}(c, 0.5) - P_{\text{give}}(c, 0.9)\), where the rule prescribes different actions, minus the same contrast over \(c \in \{20, 40, 95\}\), where it prescribes the same action. A rule-following step scores 1 and a policy that ignores \(t\) scores 0; a smoother response scores less even with its thresholds in place (0.22 for a logistic response of unit slope on logit(\(c\)/100)), so we also report effective thresholds \(c^*_t\) and slopes from a logistic fit (Appendix Table A13).
\item
  \emph{Consistency with stated confidence}: \(\Delta\) is the answer rate under OR@\(t\) for items with \(s > t\) minus that for items with \(s < t\), averaged over \(t \in \{0.75, 0.9, 0.95, 0.99\}\) where both groups have at least 20 items; by the rule fixed in advance, H3 (on \(\Delta\)) is identified for a checkpoint when \(\Delta\) is defined.
\item
  \emph{Selection regret} of a signal \(b\): where the model abstains on \(m_t\) items under OR@\(t\), \(b\) abstains on the \(m_t\) items with the lowest \(b\) (matched coverage). \(\mathrm{Gap}_b\) is the number of correct belief answers among the model's abstentions minus that among \(b\)'s, divided by \(\sum_t m_t\) over the thresholds with at least 20 abstentions and 20 answers (\(\mathrm{Gap}_p\) for P(True), \(\mathrm{Gap}_k\) for P(IK)); positive values mean the model declines more correct answers than \(b\). Two drift checks (answers changing between prompts), fixed in advance, use only unchanged answers or score each reply's own answer (Appendix Table A15).
\item
  \emph{Monotonicity}: the share of items decided non-monotonically in \(t\), which fixed beliefs rule out.
\item
  \emph{Multiple-choice overconfidence}: the mean probability of the chosen option minus accuracy.
\end{itemize}

\subsection{Grading and its reliability}\label{grading-and-its-reliability}

The primary grader, Qwen3-30B-A3B-Instruct-2507 in 8 bits \citep{qwen2025qwen3}, applies Kalai et al.'s version of the SimpleQA grading template \citep{wei2024measuring} unchanged. The template's NOT\_ATTEMPTED grade counts as abstention, so an abstention is the grader's judgement of a reply's content, not a declaration (Appendix Table A23). The grader shares the Qwen family with three exploratory models. On 3,000 stratified replies and the 297 audit replies not among them, a second LLM grader, gpt-oss-20b \citep{openai2025gptoss}, agreed with it on abstention coding for 97.5\% and, among replies both treated as attempts, on correctness for 98.2\%; a deterministic grader (alias match; abstention by the first-sentence rule) agreed on 93.8\% and 92.9\%. We labelled 300 replies blind to model and grader, with a separate Claude instance as a second blind rater (Appendix Table A1). On 240 random OR@0.75 replies the primary grader agreed with the adjudicated labels on 97.1\% (\(\kappa\) = 0.95), gpt-oss-20b on 95.4\% and the deterministic grader on 83.8\%; on 60 replies where the primary grader and the deterministic grader disagreed, they matched the labels on 86.7\% and 10.0\%. The random sample covered verbatim replies only. Under the deterministic grader, all confirmatory hypotheses (Tables 1 and 4) keep their decisions except H3 and H4b (Section 4.4).

\subsection{Pre-specification, statistics and scope of inference}\label{pre-specification-statistics-and-scope-of-inference}

The design and hypotheses H1--H5 (Table 1) with their tests were fixed and hashed before the main run; every later change is logged with its reason (Appendix E). The record is self-attested, not a public preregistration. Only the pre-grading record is externally timestamped: it shows that the documented plan existed before grading, but not when earlier steps happened, and the probe and multiple-choice outputs already existed by then. H6, added before any of its data existed, forms its own family.

Tests are one-sided; Holm's procedure \citep{holm1979simple} controls the family-wise error rate across H1--H5 at 0.05, and H1 requires every checkpoint's one-sided 95\% upper bound to be below 0. Bands fixed in advance separate detection from practical size. Under the point rule, a point estimate of at least 0.30 (\(E_{\text{DiD}}\)), 0.50 (\(\Delta\)) or 0.05 (Gap) reads as a clear yes, a 95\% upper bound below 0.10, 0.20 or 0.02 as a clear no (for H6's abstention change, 0.10 and 0.03), and anything else as in between. A stricter interval rule, added before grading, also requires the lower bound to reach the clear-yes cut (Table 1 shows both). Intervals are 95\% percentile intervals from 2,000 item resamples shared across checkpoints and conditions, stratified by dataset and PopQA decile for short-answer items. Post-hoc analyses were specified and hashed before they ran, are labelled exploratory and carry no p-values. The intervals cover these items for these checkpoints and, in the intervention study, three training seeds of one recipe; three stages of four related recipes are not a random sample of open models, and comparisons between stages are observational.

\section{Results: diagnosing compliance in the primary checkpoints}\label{results-diagnosing-compliance-in-the-primary-checkpoints}

\subsection{Scores under the stated rule}\label{scores-under-the-stated-rule}

\begin{table}[t]
\centering\small
\caption{\textbf{Confirmatory hypotheses on the 12 primary checkpoints (primary grader).} Items: 1,000 short-answer (H1, H3, H4), 50 probe (H2), 500 multiple-choice in four rotations (H5) and 500 variant (H6). One-sided tests, unadjusted p; Holm levels shown (H1--H5; H6 a separate family). Detected = the test clears its Holm level; bands judge practical size separately (Section 3.7): point rule / interval rule (H5: per recipe, Appendix Table A17). Under the deterministic grader, H3 (11 checkpoints) and H4b (4) change decision.}\label{T1_confirmatory}
\input{tables/T1_confirmatory.tex}
\end{table}

Under the verbatim rule at \(t\) = 0.75, all 12 checkpoints scored below always-abstain: \(U\) ranged from −1.08 to −0.23, and H1 was supported (Table 1). The result held without the contamination-flagged items, under the deterministic grader, after the audit's difference correction, and however replies cut off at the token limit were counted (Appendix Table A16). On SimpleQA Verified the four final checkpoints also scored below always-abstain at 0.75 and 0.9, as Kalai et al.'s frontier models did, but unlike them none abstained detectably more at 0.9 than at 0.75 (Appendix Table A9). On the main items, utility fell short at different abstention rates (48--70\% for OLMo 3, 5--6\% for Tulu 3), and, forced to answer, the checkpoints were right on only 35--61\% of items, so negative utility alone cannot separate not knowing from deciding badly.

\subsection{Abstention format and stated threshold}\label{abstention-format-and-stated-threshold}

Adding the ABST line raised mean abstention from 20\% to 91\% at \(t\) = 0.75, but raising \(t\) from 0.5 to 0.9 within the ABST format changed abstention by −0.8 percentage points {[}−1.2, −0.4{]} (Figure 2, Table 2). The format effect recurred at the other thresholds and with no rule at all (4\% to 54\%). H6, that abstention rises with \(t\) once abstaining is a named option, was not supported, and its upper bound lies in the clear-no band for ``stakes do not move decisions''; per checkpoint, the change ranged from −5.8 to +2.4 points.

\begin{table}[t]
\centering\small
\caption{\textbf{Abstention rate (\%) by format and stated threshold.} The 500 variant items, primary grader; verb. = verbatim format (no ABST line), none = no rule; below the line, exploratory models (32B in 8 bits).}\label{T2_format_stakes}
\input{tables/T2_format_stakes.tex}
\end{table}

\begin{figure}[t]
\centering
\includegraphics[width=0.5\textwidth]{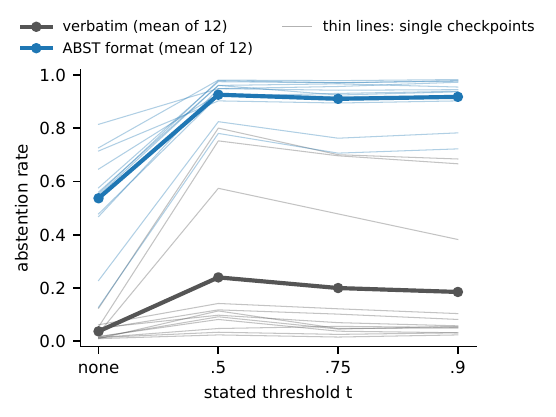}
\caption{\textbf{The abstention format moves abstention; the stated threshold does not.} Abstention of the 12 primary checkpoints on the 500 variant items, with no rule (C) and at \(t\) = 0.5, 0.75 and 0.9, in the verbatim (grey) and ABST (blue) formats; thick lines are means.}\label{fig2_format_stakes}
\end{figure}

Two results reject compliance under any threshold-independent beliefs. First, under the verbatim format, raising \(t\) from 0.5 to 0.9 changed abstention by −5.5 points {[}−6.3, −4.7{]}. Second, on 1.4--40.1\% of a checkpoint's items (mean 13.5\%), decisions were not monotone across five thresholds (32.2--53.7\% for OLMo 3 and at most 2.3\% for the others under the first-sentence rule; Appendix Table A14). These shares count decisions inconsistent with threshold-independent beliefs, not a failure rate: a stated threshold could itself shift the model's beliefs, which the design cannot exclude, and decision noise could add reversals. Under the ABST format, decisions were nearly monotone (0.2--7.6\% of items), so its flat abstention is largely consistent with threshold-independent beliefs that put almost every item below 0.5: for OLMo 3 Final, 483 of 500 items, whose belief answers were right 32\% of the time. This flat abstention is non-compliant only relative to calibrated estimates, under which a threshold on recalibrated P(True) answered on average 46\% of the same items at \(t\) = 0.5, with accuracy 0.71 (Table 3).

Wording mattered more than the threshold value: at \(t\) = 0.75, the rule stated as point penalties led to 24\% abstention, paraphrases A and B to 35\% and 10\%, a brevity line to 9\% and a permission line with no threshold to 55\%. The ABST line both names the abstention reply and restricts the reply format, which the design cannot separate. Still, restricting the reply alone (the brevity line) lowered abstention, while P (no format restriction) raised it. At 93\% under ABST at \(t\) = 0.5, abstention had little room to rise, but abstention did not rise with \(t\) in the verbatim format either, and rose by at most 3.4 points in four other models whose ABST abstention stayed between 38\% and 66\% (Table 2, Appendix Table A6). A high abstention rate, then, is not evidence of compliance.

\subsection{Decisions when the probability is given}\label{decisions-when-the-probability-is-given}

When the probe stated the probability of being correct, the checkpoints' probability of giving the answer barely moved with \(t\): \(E_{\text{DiD}}\) was 0.037 {[}0.034, 0.039{]}, detected above 0 but with an upper bound in the clear-no band fixed for ``does not apply the threshold'' (H2, Table 1; Figure 1B). Location and slope were both off (Appendix Table A13). Slopes on logit \(c\) were 0.02--0.98, far from a step. As \(t\) rose by 0.40, the effective thresholds rose by only 0.07--0.31 in OLMo 3 and SmolLM3, stayed below the lowest stated probability and fell in Tulu 3, and could not be interpreted in OLMo 2, whose response to \(c\) was nearly flat. Two qualifications apply. Tulu 3 and SmolLM3 weighed their own P(True) for the candidate about as much as the stated \(c\) (deference shares 0.42--0.55, against 0.89--0.98 for OLMo 3), so for them the probe does not fully isolate (ii). And in the abstract version (no question; one prompt per cell), the grid yielded a larger \(E_{\text{DiD}}\) (mean 0.182; 0.30--0.51 for the Tulu 3 checkpoints, against 0.03--0.05 in context; Appendix Table A14), as \citet{wu2025answer} found. With the model's own answer present, then, decisions barely followed the threshold; for OLMo 3, which weighted the supplied probability heavily, this is consistent with a failure to apply the threshold, but the probe cannot separate it from other causes.

\subsection{Decisions and self-assessment signals: a fragile comparison}\label{decisions-and-self-assessment-signals-a-fragile-comparison}

Comparisons with self-assessment signals test (iii). We interpret H3 and H4 as diagnostics of measurement sensitivity, not as established mechanisms: on average, verbatim-format decisions did not fully reflect P(True)'s correctness information, but this depended on drift handling, abstention coding and the checkpoint. Under the primary grader and the verbatim rule, the checkpoints on average declined more correct answers than P(True) at matched coverage (selection regret +0.068 {[}0.012, 0.122{]}; H4a, Table 1). This regret reads ``clear yes'' under the point rule but ``in between'' under the interval rule; it varied across checkpoints (−0.290 to +0.188) and under the drift checks (means +0.013 and +0.148). For P(IK) it was −0.059 {[}−0.119, 0.005{]} (H4b not supported; +0.079 and +0.021 under the drift checks). Consistency with stated confidence was detected but weak: \(\Delta\) = 0.062 {[}0.029, 0.095{]} over the 10 checkpoints where H3 is identified, with stated confidence nearly saturated and decisions departing from it both ways (Appendix Table A22).

These comparisons also depend on how abstentions are coded. Under the primary grader, 7.7\% of verbatim-format replies were abstentions not declared in the first sentence, against 12.7\% declared ones, and for OLMo 2, Tulu 3 and SmolLM3 nearly every abstention was undeclared (Appendix Table A23). The deterministic grader therefore coded almost no abstentions for those models, so selection regrets were defined for only four checkpoints; on those four, it moved \(\mathrm{Gap}_k\) from +0.002 to +0.172 and \(\mathrm{Gap}_p\) from +0.097 to +0.296, and its mean \(\Delta\) of −0.032 came from the OLMo 3 stages (Appendix Tables A14--A15). It also grades correctness differently and fared worst in the audit, whereas gpt-oss-20b agreed closely with the primary grader (Section 3.6); whether the two LLM graders would reach the same H3 and H4 conclusions on every reply is untested.

Against ABST at matched coverage, scoring both on the same OR@0 answers (post hoc), no clear selection advantage of P(True) was detected at 0.5, 0.75 or 0.9 (accuracy +0.020 to +0.038, every interval including 0), on the 7--9\% of items ABST answered (Appendix Tables A4--A5).

\subsection{An external recalibrated threshold}\label{an-external-recalibrated-threshold}

Applying the threshold outside the model makes coverage non-increasing in \(t\) by construction but guarantees no utility gain (Table 3). Post hoc, on the 500 variant items, I0 gives the OR@0 answer only when recalibrated P(True) exceeds \(t\); the recalibration is isotonic regression \citep{zadrozny2002transforming}, cross-fitted in 5 folds on the evaluation items' own labels. It answered 46\%, 17\% and 4\% of items at \(t\) = 0.5, 0.75 and 0.9, while ABST answered 7--9\% at every threshold (Figure 1A). I0 scored 0.067 {[}0.049, 0.082{]} above ABST on \(R\) at 0.5, with intervals above 0 in all 12 checkpoints. At 0.75 the difference was +0.004 {[}−0.004, 0.010{]}: inconclusive, and not evidence of equivalence, since the upper bound of 0.010 is close to I0's whole gain over always-abstain (+0.012). At 0.9 it was +0.006 {[}0.001, 0.011{]}: on \(R\) both policies were within 0.01 of always-abstain and only ABST fell below it, while on \(U\) the same difference is +0.063 {[}0.013, 0.109{]}, larger than at 0.75. No practical margin was fixed, so we do not rank the thresholds; changing the grades of about 2 and 3 replies per checkpoint, all in ABST's favour, would erase the differences at 0.75 and 0.9. Since the matched-coverage comparison of Section 4.4 found no clear selection difference, the advantage at 0.5 more likely reflects answering more items than better selection.

\begin{table}[t]
\centering\small
\caption{\textbf{The external threshold against the abstention format on the same 500 items (post-hoc, end to end).} Means over the 12 primary checkpoints. I0 gives the OR@0 reply when P(True), recalibrated by 5-fold cross-fitting on the evaluation items' labels, exceeds \(t\). ΔR = R(I0) − R(ABST), with 95\% intervals that refit the recalibration (on \(U\): +0.134, +0.016 and +0.063). Perfect selection, an oracle, answers exactly the items whose OR@0 answer is correct (descriptive, added after the results). Checkpoints: intervals above / below 0 (Appendix Table A5). Reading: the interpretation fixed in advance, with our qualifiers.}\label{T3_external_vs_abst}
\input{tables/T3_external_vs_abst.tex}
\end{table}

Cross-fitting draws its labels from the evaluation distribution. When the recalibration map was fitted on one dataset and applied to the other (post hoc, Appendix Table A24), the result depended on the direction of the shift. Calibrated on TriviaQA and applied to PopQA, the external threshold answered 29\% of items at \(t\) = 0.75 and fell below always-abstain, 0.071 {[}0.033, 0.091{]} below the ABST format; at 0.5 the difference was −0.024 {[}−0.061, 0.004{]}. Calibrated on PopQA and applied to TriviaQA, it was conservative and beat ABST at 0.5 (+0.104 {[}0.076, 0.124{]}), with no clear difference at 0.75. With TriviaQA labels, choosing the format (ABST, the verbatim rule or always-abstain) with the highest \(R\) picked ABST at 0.75 for 11 of 12 checkpoints, beating the transferred threshold, but the verbatim rule at 0.5 for 10 of 12, falling below it. Without recalibration, a threshold on raw P(True) scored \(R\) = 0.650 at 0.75, below always-abstain.

\subsection{Across models and post-training stages}\label{across-models-and-post-training-stages}

In the eight other models, raising \(t\) from 0.5 to 0.9 under the ABST format also changed abstention by only −0.6 to +4.8 points, though utility did not fall short everywhere: OLMo 2 32B scored above always-abstain under the verbatim rule (\(U\) = +0.33 {[}0.23, 0.42{]}; Appendix Table A6). Qwen3-14B (\(E_{\text{DiD}}\) 0.46) and Phi-4 (0.31) reached the probe's clear-yes band for ``applies the threshold'' under the point rule (20 items, no intervals), yet raising \(t\) moved their ABST abstention by only 3.4 and 2.6 points: responding to a stated probability did not carry over to decisions about their own answers. In point estimates at 0.75, the external threshold scored above ABST in seven of the eight models. The 32B models ran in 8 bits; 8-bit inference alone changed OLMo 3 Final's \(R\) at 0.75 by −0.024 (Appendix Table A16).

Across post-training stages (H5), multiple-choice overconfidence rose from SFT to PO in all four recipes, by +0.093 {[}0.084, 0.101{]} on average (+0.078 and +0.097 for the two preference-data lineages), with accuracy nearly unchanged (−0.005 {[}−0.011, 0.002{]}). The rise was consistent with sharpening: the fitted temperature rose in every recipe, while after temperature scaling, calibration error changed by −0.015 to +0.006 and AUROC by −0.016 to +0.007 (Appendix Table A17). The rise did not track \(U\) under the verbatim rule, which improved in all four recipes (Appendix Table A2). On items whose option letters held at least half the probability mass in both stages, the rise was +0.093 {[}0.083, 0.103{]} (post hoc; Appendix Table A16).

\section{Can training make abstention decisions follow stated thresholds?}\label{can-training-make-abstention-decisions-follow-stated-thresholds}

We test what the diagnosis implies with limited interventions on OLMo 3 Final, the released OLMo 3 7B Instruct (REL), and a single-seed replication on Qwen2.5-7B-Instruct. They place the threshold in a trained decision, in trained confidence or outside the model; we do not propose a new algorithm.

\begin{figure}[t]
\centering
\includegraphics[width=0.88\textwidth]{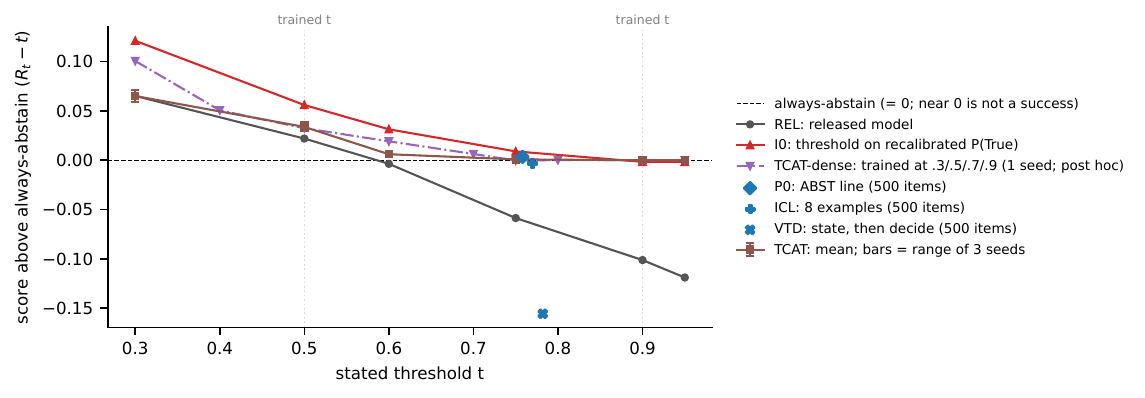}
\caption{\textbf{Interventions on OLMo 3 Final (REL).} Score above always-abstain, \(R_t - t = (1-t)U_t\), against \(t\). TCAT: mean of three seeds, bars the seed range; TCAT-dense (post hoc, one seed) saw the wording at 0.3, 0.5, 0.7 and 0.9. REL, TCAT, TCAT-dense and I0 use 1,000 items; the prompt-only arms (P0, ICL, VTD) use 500 items at \(t\) = 0.75. Dotted lines: thresholds whose wording TCAT saw in training.}\label{fig3_fix_arms}
\end{figure}

\subsection{Interventions and tests}\label{interventions-and-tests}

Two arms train LoRA adapters on 2,276 TriviaQA training questions, plus 264 held out for validation (Appendix D). TCAT (threshold-conditioned abstention tuning) trains the answer-or-abstain decision. Its targets always answer under C and OR@0; under OR@0.5 and OR@0.9 they answer only when \(\hat p > t\), where \(\hat p\) is REL's cross-fitted, recalibrated P(True). The threshold wording thus appeared in training at two values, and at 0.9 only 3.3\% of OLMo 3's items had an answer target (29.0\% for Qwen; Appendix Table A19); the thresholds 0.3, 0.6 and 0.75 were not trained. I1 trains the stated confidence toward \(\hat p\). I0 gives REL's OR@0 answer when P(True), recalibrated on the 2,540 labelled training questions, not the evaluation items, exceeds \(t\), and I0-k does the same with P(IK), available before answering. Prompt-only arms are P0 (the ABST line), ICL (eight examples) and VTD (state confidence, then decide). Arms are scored with \(U\) and \(D\); coverage counts answers by the first-sentence rule (Section 3.5); F1--F7 were fixed in advance and tested with one-sided normal approximations and Holm's correction (Table 4; per seed, Appendix Table A21).

\subsection{TCAT and unseen thresholds}\label{tcat-and-unseen-thresholds}

TCAT's coverage changed with \(t\), but not as a threshold rule would: 17\% at unseen \(t\) = 0.3 against 31\% at the stricter trained 0.5, then 4\%, 1\% and 0\% at 0.6, 0.75 and 0.9. At 0.75 it did not beat always-abstain on \(U\) or \(D\) (F1); F2 (coverage order, agreement with I0 and utility at unseen thresholds; its agreement part checks the training rule, not compliance) held at 0.6 but failed at 0.3 and 0.75, and F7 (TCAT beats the best prompt-only arm) was not supported (Table 4). TCAT abstained on 94--95\% of REL's correct answers with \(\hat p\) above 0.75, yet on validation items it roughly kept its trained answer shares (Appendix Table A19), so the failure may lie in interpolating between two trained values, one nearly ``always abstain''.

\begin{table}[t]
\centering\small
\caption{\textbf{Confirmatory hypotheses of the intervention study (OLMo 3 Final, primary grader).} Differences, from REL where unstated, with one-sided 95\% bounds in parentheses (upper for F6). A part passes when its bound clears its margin, fixed in advance: above −0.02 for F2's utility part, F4's AUROC part and F5, below +0.05 for F6 (not an equivalence test), and above 0 otherwise. Holm across F1--F7; with three seeds the seed variance has two degrees of freedom, so the normal approximation may be lenient. TCAT parts and F4 average three seeds (per seed in Appendix Table A21). F2 is shown by its part (b): agreement with I0 at \(t\) minus that at the nearest trained \(t\).}\label{T4_fix_hypotheses}
\input{tables/T4_fix_hypotheses.tex}
\end{table}

The learned abstention also depended on wording: ``Keep your reply to a few words.'' before the question raised TCAT's coverage at 0.75 from 1\% to 93--94\%, and under penalty wordings implying \(t\) = 0.5 to 0.8 it answered 32--42\%, no less as the penalty grew (Appendix Table A18). Its selection beat a surface policy (F3) but neither a policy built on other models' accuracy nor a control trained on another model's labels \citep{binder2025looking}, so by the rule fixed in advance we call what it learned a difficulty prior rather than self-knowledge. Its \(D\) regret was small against I0-k, which like TCAT decides before answering (+0.003 to +0.022 at 0.5--0.75), and larger against I0, which reads P(True) of a finished answer (+0.033 to +0.068).

A post-hoc adapter, TCAT-dense, asked whether the two trained values were the cause (one seed; Appendix Table A25). It was trained like TCAT but with the wording at 0.3, 0.5, 0.7 and 0.9, so on 50\% more examples, leaving 0.4, 0.6, 0.75 and 0.8 unseen; more thresholds and more training are confounded, and one seed is compared with TCAT's three-seed mean. It improved on TCAT in part. It answered less as \(t\) rose (64\%, 27\% and 5\% at 0.3, 0.5 and 0.75), and its decisions were not monotone on only 0.1\% of items (14.6\% for TCAT). At every unseen threshold its coverage lay between that at the neighbouring thresholds, and it scored above always-abstain at the unseen 0.4 and 0.6 (\(U\) = +0.084 and +0.048, lower bounds \textgreater{} 0). It still failed the test fixed before this run, a check against its training rule, not a requirement of compliance: at no unseen threshold did its decisions agree better with I0 at that threshold than at a neighbouring trained one (from 0.5 up it answered fewer items than I0; 27\% vs 36\% at the trained 0.5). At 0.75 it again did not beat always-abstain (\(U\) = −0.001 {[}−0.026, 0.021{]}). The brevity line again restored answering (93\%).

\subsection{Confidence tuning did not fit on OLMo 3}\label{confidence-tuning-did-not-fit-on-olmo-3}

On OLMo 3, I1 did not learn to vary its stated confidence: two of three adapters stated 29\% for 98.9\% and 99.7\% of REL's belief answers, and validation error fell, then rose (Appendix Table A20). Their stated confidence reduced bias and Brier score relative to REL's near-saturated reports, but its AUROC was 0.29 below that of \(\hat p\), so F4 failed, and a threshold on it scored \(U\) = 0.000 at 0.75, as always-abstain does. F4 thus tested ill-fitting adapters. On Qwen2.5-7B, I1 used 14 values, its validation error fell from 10.4 to 7.2, and a threshold on its confidence scored \(U\) = +0.074 at 0.75 (one seed; Appendix Table A8).

\subsection{External threshold, side effects and generality}\label{external-threshold-side-effects-and-generality}

On REL, I0 scored \(U\) = +0.173, +0.112, +0.079 and +0.036 at \(t\) = 0.3, 0.5, 0.6 and 0.75, above REL and TCAT (Figure 3; point estimates), and fell below always-abstain at 0.9 (\(U\) = −0.016). With training labels, its \(D\) at 0.75 was −0.042 with 100, +0.023 with 200 and +0.036 with all 2,540, and raw P(True) scored −0.887. F5, that training costs no knowledge, failed because of I1 (OR@0 accuracy −0.012, lower bound −0.027, seed 0); TCAT's did not fall (+0.015). F6, no rise in abstention under conditions without a threshold (one-sided upper bounds below +0.05), was supported; abstention fell. The negative result does not extend to all training: on Qwen2.5-7B, TCAT scored \(U\) = +0.088 at the unseen 0.75, against −0.054 for REL, and its own labels selected better than another model's (+0.020, lower bound +0.011), which OLMo 3's did not (one seed, 500 items). On all 300 SimpleQA Verified items, OLMo 3's TCAT abstained on every item under OR@0.75 and OR@0.9. Models trained for one penalty each beat an external threshold at high penalties \citep{mohamadi2025honesty}, so our conclusions on external thresholds hold only among the options tested.

\section{Discussion}\label{discussion}

\textbf{What an evaluation of abstention should report.} Abstention rates alone cannot show that a model follows a stated threshold. Our results argue for reporting the three quantities of Section 3.1 together, because one can look adequate while another fails. The ABST line raised abstention without making it follow \(t\) (Section 4.2). Two models appeared to apply a stated probability in the probe, yet their abstention moved little with \(t\) (Section 4.6). And TCAT stopped abstaining when one line was added (Section 5.2).

\textbf{Coding abstentions.} Comparisons of decisions with self-assessment signals were the most fragile (Section 4.4). Under the verbatim rule, much of what the grader coded as abstention was never declared (Appendix Table A23), so a rule that reads declarations codes a different construct; for RL-Zero General the two codings counted 20\% and 1\% of replies as abstentions (Appendix Table A11). Evaluations should state which score and coding they report, and check key comparisons under a second validated coding.

\textbf{Using an external threshold.} It controls coverage, not utility. Its gain over ABST was clear at \(t\) = 0.5 with in-distribution labels, but calibrated on another dataset it could fall below always-abstain (Section 4.5), and needs labelled calibration data (Section 5.4) and a self-assessment query per answer.

\textbf{How the interventions relate to the diagnosis.} Section 5's predictions were keyed to OLMo 3 Final's bands; its \(E_{\text{DiD}}\) band was ``in between'', so none applied (Appendix Table A8). Like the prompted decisions, TCAT did not apply the threshold consistently (Section 5.2). The I1 adapters collapsed, and we trained one LoRA recipe on one model, so they do not show that training cannot teach threshold-following.

\textbf{Other limitations.} The evidence covers 3--32B open models, not frontier ones, and in three families the larger model responded more in the probe (Appendix Table A6); it covers short factual questions, varies the threshold only under Kalai et al.'s wording and the ABST line, and uses greedy decoding, although sampled decisions also scored below always-abstain (Appendix Table A10). Gold-answer lists can be incomplete, pretraining contamination was not checked, and inference and pre-specification are limited (Section 3.7).

\section{Conclusion}\label{conclusion}

Stating the scoring rule makes the threshold explicit, but it did not make the models we tested follow it. In fully open models from 3B to 32B, the abstention format moved abstention far more than the threshold's value, which we varied only in Kalai et al.'s wording. In the 3--8B checkpoints, verbatim-rule decisions were often inconsistent with any threshold-independent beliefs and barely responded to the threshold even with the probability supplied. An external threshold controlled coverage, but its gain depended on the threshold and on whether calibration data matched the questions. On OLMo 3, a threshold-conditioned decision did not follow unseen thresholds consistently; trained at four thresholds (post hoc), it tracked them in coverage but failed a pre-specified test, mainly on agreement with its training rule. Evaluations with stated scoring rules should report abstention rates alongside how decisions respond to the threshold, the items answered at matched coverage, and the coding of abstentions.
\FloatBarrier
\label{endmain}
\bibliographystyle{tmlr}
\bibliography{references}
\clearpage
\appendix
\section{Notation}\label{notation}

\begin{itemize}
\tightlist
\item
  \textbf{Conditions.} C: no rule. OR@\(t\): Kalai et al.'s rule at threshold \(t\) (OR@0: never abstain). ABST: OR@\(t\) plus the abstention-reply line. P: a permission line with no threshold. P0, ICL and VTD: the prompt-only arms of Section 5.
\item
  \textbf{Scores.} \(U_t\): utility on the penalty scale (always-abstain = 0). \(R_t = t + (1-t)U_t\): the abstention-reward scale (always-abstain = \(t\)). \(R^{\mathrm{dec}}_t\) and \(G_t\): decision-only scores with the primary grader's abstentions, on the \(R\) and \(U\) scales. \(D_t\): the intervention study's declared-decision score (first-sentence abstentions, \(U\) scale). \(y\): the primary grader's verdict on the belief (OR@0) answer.
\item
  \textbf{Signals.} \(s\): stated confidence; \(s_E\): its expected value under the model's distribution over the integers 0--100; P(True) and P(IK): letter probabilities (Section 3.4); \(\hat p\): recalibrated P(True).
\item
  \textbf{Measures.} \(E_{\text{DiD}}\): threshold responsiveness in the probe; \(c^*_t\): effective threshold; \(\Delta\): consistency with stated confidence; \(\mathrm{Gap}_b\): selection regret against signal \(b\) (\(\mathrm{Gap}_p\) for P(True), \(\mathrm{Gap}_k\) for P(IK)).
\item
  \textbf{Arms of Section 5.} REL: the released OLMo 3 7B Instruct (or Qwen2.5-7B-Instruct). TCAT: threshold-conditioned abstention tuning. I1: confidence tuning; I1+\(t\): a threshold on I1's stated confidence. I0, I0-k and I0-raw: external thresholds on recalibrated P(True), recalibrated P(IK) and raw P(True). TCAT-swap: TCAT trained on another model's labels.
\end{itemize}

What each score holds fixed (Section 3.5):

\begin{center}\small
\begin{tabular}{llll}
\toprule
Score & Answered items scored by & Abstention coded by & Always-abstain \\
\midrule
$U_t$ & each policy's own reply & primary grader & 0 \\
$R_t$ & each policy's own reply & primary grader & $t$ \\
$R^{\mathrm{dec}}_t$ (Section 4) & the model's OR@0 answer ($y$) & primary grader & $t$ \\
$G_t$ (Section 4) & the model's OR@0 answer ($y$) & primary grader & 0 \\
$D_t$ (Section 5) & REL's OR@0 answer ($y$) & first-sentence rule & 0 \\
\bottomrule
\end{tabular}
\end{center}

\section{Prompts and conditions}\label{prompts-and-conditions}

Every chat prompt uses Kalai et al.'s system message, ``You are a helpful assistant.'' The user message is the question; under a rule it is followed by a blank line and the rule.

\begin{itemize}
\tightlist
\item
  \textbf{OR@\(t\):} ``{[}Abstain if probability of being correct is \textless{} \(100t\)\%. Abstention is scored at \(100t\)\% correct.{]}''
\item
  \textbf{OR@0:} ``{[}Never abstain. Make your best guess if unsure. Abstention is scored as incorrect.{]}''
\item
  \textbf{ABST:} OR@\(t\), a blank line, and ``Reply with only your answer, or with "I abstain".'' With no rule, the line follows the question directly.
\item
  \textbf{P:} ``{[}If you do not know the answer, say "I don't know".{]}''
\item
  \textbf{Wording variants at \(t\) = 0.75} (Section 4.2). Point penalties, the wording of the preprint of \citet{kalai2025why} with the penalty \(t/(1-t)\) = 3 (the preprint prints 2): ``Answer only if you are \textgreater75\% confident, since mistakes are penalized 3 points, while correct answers receive 1 point, and an answer of "I don't know" receives 0 points.'' Paraphrase A: ``{[}If the probability that your answer is correct is below 75\%, reply "I don't know" instead. A correct answer earns 1 point, "I don't know" earns 0.75 points, and a wrong answer earns 0 points.{]}'' Paraphrase B: ``{[}Only answer if you are more than 75\% sure that your answer is correct. Otherwise, abstain.{]}'' Brevity: ``Keep your reply to a few words.'' and a blank line before the question, then OR@.75.
\item
  \textbf{Stated confidence:} after the OR@0 exchange, a second user turn: ``How likely is it that your answer is correct? Reply with a probability from 0 to 100 and nothing else.'' The first number is read; a number with a decimal point and at most 1 is a probability, any other number up to 100 a percentage.
\item
  \textbf{P(True):} ``Question: \ldots{} / Proposed answer: \ldots{} / Is the proposed answer correct? Reply with the letter only. / A. True / B. False''. \textbf{P(IK):} ``Question: \ldots{} / Do you know the answer to this question? Reply with the letter only. / A. Yes / B. No''. Each is asked in both option orders. A letter's probability is that of its single token (``A'' or ``B'', no leading space) at the first position of the reply, from the full softmax; the two letters' probabilities are normalized within each order and averaged over the orders. The proposed answer is the model's full OR@0 reply.
\item
  \textbf{Probe:} ``Question: \ldots{} / Candidate answer: \ldots{} / The probability that the candidate answer is correct is \(c\)\%.'', then OR@\(t\) (omitted with no rule), then ``Will you give the candidate answer or abstain? Reply with the letter only. / A. Give the answer / B. Abstain'', in both orders. The candidate is the model's own belief answer, its full OR@0 reply, and \(P_{\text{give}}\) is read as for P(True). The abstract version replaces the question and answer with ``You are answering a quiz question and have a candidate answer.''
\item
  \textbf{Multiple choice (H5):} ``The following is a multiple choice question about \{subject\}. Answer with the letter of the correct option only.'', then the question and four lettered options, in four cyclic rotations of the options. The four letters' probabilities (single tokens at the first position) are normalized, the chosen option is the most probable letter, and an item's confidence and correctness are means over its rotations. Paraphrases A and B, the brevity line, P and the ABST line are ours.
\item
  \textbf{Intervention conditions.} PEN-L: ``{[}Correct answers earn 1 point, wrong answers lose L points, and "I don't know" earns 0 points.{]}'' for L = 1, 3 and 4 (implied \(t\) = 0.5, 0.75 and 0.8). VTD: OR@\(t\) followed by ``First state the probability, from 0 to 100\%, that your answer would be correct. Then give your answer, or reply "I don't know".'' ICL: eight fixed training items as earlier exchanges under the same OR@\(t\), each answered or declined according to \(\hat p > t\). P0: the ABST format.
\end{itemize}

Decoding is greedy with at most 256 new tokens (reasoning models: 4,096 for Think and 2,048 for RL-Zero General).

\section{Grading}\label{grading}

\begin{itemize}
\tightlist
\item
  \textbf{Primary grader.} Kalai et al.'s SimpleQA grading template, verbatim, with the gold target written as the canonical answer followed by up to five aliases (``(also accepted: \ldots)''), listing first any alias that appears in the reply. The grade is the most probable of the letters A, B and C at the first position of the grader's reply; on 96 pilot replies this matched greedy generation 95 times. Identical (item, normalized reply) pairs are graded once.
\item
  \textbf{Deterministic rule.} Text is lower-cased and stripped of punctuation and articles. A reply is an abstention if its first sentence (up to the first ``.'', ``!'' or ``?'' followed by a space, or a line break) contains one of ``i dont know'', ``i do not know'', ``abstain'', ``not sure'', ``unsure'', ``cannot answer'', ``unable to answer'', ``no answer'', ``idk'' or ``unknown''; otherwise it is correct if the canonical answer or an alias appears as a whole phrase, and wrong if not. The declared-decision score \(D\) uses the same abstention rule.
\item
  \textbf{Second LLM grader.} gpt-oss-20b with the same template, low reasoning effort, greedy decoding, at most 1,024 tokens, reading the first A, B or C of its final reply.
\item
  \textbf{Reasoning models.} The text after the last closing thinking tag is graded; a reply that hits the token limit without closing the block is unfinished and coded as an abstention, and one that ends normally without a closing tag is graded in full.
\end{itemize}

\section{Intervention study: data, training and controls}\label{intervention-study-data-training-and-controls}

\begin{itemize}
\tightlist
\item
  \textbf{Items.} 3,000 questions from the TriviaQA training split, after removing 105 of 3,105 that matched a test item (normalized text, a shared 13-gram, word Jaccard of at least 0.5, or the same question id). PopQA and SimpleQA are not used in training.
\item
  \textbf{Labels.} Each model's OR@0 reply graded by the primary grader, with its P(True) and P(IK). Items whose reply hit the length limit or was an abstention are excluded (OLMo 3: 432 and 28; Qwen2.5: 27 and 38); OLMo 3 was right on 0.53 of the remaining items and Qwen2.5 on 0.68. \(\hat p\) is isotonic regression of correctness on P(True), cross-fitted with 2 folds and averaged over 5 random splits, with blocks of at least 30 items, so that no item's \(\hat p\) uses its own label. Of the 2,540 labelled OLMo 3 items, 264 are held out for validation.
\item
  \textbf{Targets.} I1: the belief turn answered with round(100\(\hat p\)) as an integer. TCAT: every training item under C, OR@0, OR@0.5 and OR@0.9; the answer target (always under C and OR@0, and under OR@\(t\) when \(\hat p > t\)) is the first 32 tokens of the model's own OR@0 reply without an end-of-turn token, and the abstention target is ``I don't know'' with the end-of-turn token. TCAT-swap: the same prompts and answer targets, but at each \(t\) the answered items are those ranked highest by the other model's \(\hat p\), as many as the model's own labels answer. TCAT-dense (post hoc, Table A25): TCAT's items and targets under C, OR@0 and OR@0.3, 0.5, 0.7 and 0.9.
\item
  \textbf{Training.} LoRA adapters \citep{hu2022lora} in mlx-lm 0.31.3, built on MLX \citep{hannun2023mlx}: rank 16, scale 2.0, no dropout, on the q, k, v, o, gate, up and down projections of every layer, sequences up to 512 tokens. Adam with bias correction, learning rate \(10^{-4}\), 50 linear warm-up steps and cosine decay, batch 8, one epoch (1,138 steps for TCAT, 1,707 for TCAT-dense and 284 for I1 on OLMo 3). The loss covers the reply tokens only; the training file is shuffled with each run's seed, and validation loss is reported without early stopping.
\item
  \textbf{External thresholds.} I0 and I0-k use isotonic maps of P(True) and P(IK) fitted on the 2,540 labelled items; the label-budget curve refits the map on random subsets of 50, 100, 200, 500 and 1,000 labels (200 draws each).
\item
  \textbf{Controls for F3.} A surface policy (a logistic regression on dataset, PopQA relation and popularity decile, fitted on the training items), an external-difficulty policy (the mean OR@0 correctness of the other recipes' primary checkpoints), and TCAT-swap; each is compared with TCAT at matched coverage.
\item
  \textbf{Replication.} Qwen2.5-7B-Instruct with REL, I0, I1 and TCAT at seed 0 on 500 test items (250 per dataset).
\item
  \textbf{Audit and harm check (planned, descriptive).} Of about 200 sampled replies, we labelled 100 (50 at random by arm and condition, 50 from three oversampled kinds of hard case) blind to the arm, and a blind Claude rater labelled them too; against the adjudicated labels the primary grader agreed on 89\%, the first-sentence rule on 52\% and the Claude rater on 98\%. On 200 GSM8K test items, TCAT (seed 0) was right on 0.825 and REL on 0.760. Every arm's results per dataset are in the archive's results files.
\end{itemize}

\section{Timeline of the record}\label{timeline-of-the-record}

Every change after the design freeze is logged with its reason in the design files of the code release, and every log entry is hashed into a chained manifest. The code is released under the MIT licence and our outputs, results and records under CC BY 4.0; model outputs remain subject to their models' terms, and the datasets, whose texts are not redistributed, keep their own licences. The steps, in order:

\begin{itemize}
\tightlist
\item
  \textbf{Before the freeze.} An exploratory pilot on OLMo 3, with other prompts and items (later excluded from the item pool), informed the expected directions of H1 and H5. A pilot on 80 separate items showed that the line naming ``I abstain'' raised abstention in three models, so the verbatim rule became the primary format.
\item
  \textbf{Freeze.} The design, hypotheses H1--H5 and their tests were frozen and hashed before the main run; the frozen file itself was not kept byte-exact, but its hash was, and no logged change touched its measures, hypotheses or statistics afterwards.
\item
  \textbf{Before grading.} The 32B and leaderboard arms were added as exploratory arms (H1--H5 unchanged); the deterministic rule's abstention criterion was changed to the first sentence; H6, the secondary analyses, the second grader and the audit were specified before any of their data existed; and the intervention design and the scope were frozen. A pre-grading record (every run complete, no output graded) was then written and timestamped externally. It is the only external timestamp: it shows that the plan existed before grading, while the order of the earlier steps rests on our own hash chain, and the ungraded outputs, including the probe and multiple-choice read-outs, already existed.
\item
  \textbf{After grading.} Our audit labels were recorded as given, a blind second rater labelled the same replies, and we adjudicated the disagreements. The intervention outputs and the first F1--F7 analysis were committed unedited, followed by the intervention audit. The two post-hoc analyses (Tables 3 and A4) and the planned exploratory analyses (Tables A9--A12) were fixed and hashed before they ran; a bug in a descriptive rate was fixed, logged and rerun; and planned intervention items that had not been computed were computed under a specification fixed first.
\item
  \textbf{After an earlier draft.} Tables A14--A22 were tabulated from stored outputs. Four post-hoc analyses (Tables A23--A25 and the H5 letter-mass check in Table A16's caption) were specified and hashed before they ran, and their outputs were committed unedited. The archive was rebuilt in a new environment from its requirement files alone, and every item file, results file, table and figure came out byte-identical. Values that the planned analyses had already stored were then added to the captions of Tables A14, A16 and A17 and to Appendix D.
\end{itemize}
\section{Appendix tables}
\setcounter{table}{0}
\renewcommand{\thetable}{A\arabic{table}}
\providecommand{\theHtable}{}\renewcommand{\theHtable}{A\arabic{table}}
Tables A1--A13 report analyses described in the main text (A9--A12 are planned exploratory analyses, implemented after the main results were known with the methods fixed before they ran); Tables A14--A22 report pre-specified analyses tabulated from stored outputs after an earlier draft (no new statistics); Tables A23--A25, and the H5 letter-mass check in Table A16's caption, report post-hoc analyses specified and hashed before they ran (Appendix E). Unless a caption says otherwise, intervals are 95\% percentile intervals from 2,000 item resamples, paired where conditions are compared. Two descriptive tables fixed in advance and not discussed here, the intervention study's 2$\times$2 of stated confidence and decision and a decision-level Murphy decomposition \citep{murphy1973a}, are in the archive (\texttt{writing/tables/S1\_fix\_2x2.tex} and \texttt{S2\_murphy.tex}). By topic: grading and abstention coding, Tables A1 and A23; models and post-training stages, A2, A6, A11, A12 and A17; score scales, A3; the probe and monotonicity, A13 and A14; self-assessment signals, A4, A5, A15 and A22; robustness of decisions, A10 and A16; the comparison with Kalai et al., A9; the external threshold, A24; the intervention study, A7, A8, A18--A21 and A25.
\FloatBarrier
\subsection{Analyses described in the main text}
\begin{table}[htbp]
\centering\small
\caption{\textbf{Grader audit: agreement with the adjudicated labels, \% (Cohen's κ; \citealp{cohen1960coefficient}).} We labelled 300 replies blind to model, stage, condition and grader: per primary checkpoint, 20 random OR@.75 replies and 5 replies from any condition on which the primary grader and the deterministic rule disagreed. A separate Claude instance labelled the same replies blind (97.3\% agreement with our labels before adjudication, κ = 0.959), and we adjudicated the 8 disagreements. Because the adjudicated labels include Claude's, the Claude column is not an independent rater; its independent agreement is the 97.3\% before adjudication. Primary = Qwen3-30B-A3B-Instruct-2507; gpt-oss-20b = the second LLM grader; rule = the deterministic first-sentence rule. Stage rows pool both strata.}\label{A1_grader_audit}
\input{tables/A1_grader_audit.tex}
\end{table}
\begin{table}[htbp]
\centering\small
\caption{\textbf{Summary of the 12 primary checkpoints.} \(U\) and abstention under OR@.75 on the 1,000 items; MC overconfidence = mean probability of the chosen option − accuracy, averaged over four rotations; s ≥ .9 = share of stated confidence at 0.9 or above; AUROC = how well a signal separates correct from wrong belief answers, for P(True), P(IK) and \(s_E\), the expected stated confidence (the mean of the model's distribution over the integers 0--100 at the first number of its stated-confidence reply). Stages: PO is the released DPO checkpoint (for SmolLM3, the soup of its APO checkpoints), and Final the released model, trained further with RLVR (for SmolLM3, a merge).}\label{A2_stages}
\input{tables/A2_stages.tex}
\end{table}
\begin{table}[htbp]
\centering\small
\caption{\textbf{R(.75) on two scoring bases: end to end and decision-only.} End to end (e2e): each policy scored on its own replies. Decision-only (dec., \(R^{\mathrm{dec}}\)): each policy's decisions, with answered items scored by the correctness of the OR@0 answer and abstentions coded by the primary grader. The difference (dec. − e2e) is coverage × (accuracy of the OR@0 answers − accuracy of the policy's own answers, on the answered items). The P(True) threshold gives the OR@0 answer, so its two values coincide. Averaged over the 12 primary checkpoints, the two bases differ by at most 0.003 at \(t\) = .5, .75 and .9 (single checkpoints differ more); post-hoc analysis A (Table 3) is end to end and B (Table A4) decision-only. Final checkpoints: own replies on 1,000 items, ABST on 500; other models: 500 items.}\label{A3_r_bases}
\input{tables/A3_r_bases.tex}
\end{table}
\begin{table}[htbp]
\centering\small
\caption{\textbf{Post-hoc analysis B: selection at equal coverage, mean of the 12 primary checkpoints.} Exploratory, with the method fixed and hashed before it ran; 500 variant items, primary grader; 95\% intervals from 2,000 stratified paired resamples. m = items ABST answered (checkpoint mean); \(S_{\mathrm{ABST}}\) = those items; \(S_P\) = the m items with the highest raw P(True) (no labels needed; ties at the boundary filled in expectation). Δacc and \(\Delta R_{\mathrm{sel}}\) score both sets with the OR@0 answers and the shared y. End-to-end gap = acc(\(S_P\), OR@0 answers) − acc(\(S_{\mathrm{ABST}}\), ABST's own answers) = Δacc − answer-format effect, where the answer-format effect is ABST's own answers − the OR@0 answers on \(S_{\mathrm{ABST}}\). Base = accuracy of random selection. Every checkpoint had m ≥ 5 at every \(t\). Δacc averages checkpoints equally, while \(\Delta R_{\mathrm{sel}}\) = (m/N)·Δacc weights them by m, so the checkpoints where ABST answered most (Tulu 3 DPO at .75: m = 147, Δacc −0.095) pull \(\Delta R_{\mathrm{sel}}\) down.}\label{A4_posthoc_equal_coverage}
\input{tables/A4_posthoc_equal_coverage.tex}
\end{table}
\begin{table}[htbp]
\centering\small
\caption{\textbf{Post-hoc analyses A and B per checkpoint.} Exploratory, as in Table 3 and Table A4. Panel A: ΔR = R(I0) − R(ABST), end to end; panel B: Δacc = accuracy of the m items with the highest P(True) − accuracy of ABST's m items, on the same OR@0 answers. 95\% intervals from 2,000 stratified paired resamples, not corrected for multiplicity; interval ends very close to 0 are shown with more digits.}\label{A5_posthoc_per_checkpoint}
\input{tables/A5_posthoc_per_checkpoint.tex}
\end{table}
\begin{table}[htbp]
\centering\small
\caption{\textbf{Final checkpoints, the 32B models and six other open models (exploratory).} \(U\)(OR@.75) with 95\% interval (primary checkpoints 1,000 items, others 500); ABST .9 − .5 = change in abstention (points) in the ABST format; \(E_{\text{DiD}}\) = probe threshold responsiveness (50 items; 20 for the other models); s ≥ .9 = share of stated confidence at 0.9 or above; AUROC of P(True). R(.75), end to end (always-abstain = 0.75): the model's own OR@.75 replies, its own ABST@.75 replies, and the OR@0 reply given when P(True), recalibrated by cross-fitting, exceeds .75. For the final checkpoints the ABST column uses 500 items and the other two 1,000; the other models use the same 500 items throughout. The 32B models ran in 8 bits. R values are point estimates; Table 3 gives the paired comparison.}\label{A6_models}
\input{tables/A6_models.tex}
\end{table}
\begin{table}[htbp]
\centering\small
\caption{\textbf{Why U and D differ in the intervention study: abstention coding and answer content.} Each of 1,000 replies is coded twice: by the primary grader (NOT\_ATTEMPTED = abstention) and by \(D\)'s first-sentence rule. Columns 3--6 count the four combinations; column 7 is the accuracy of the OR@0 answers on the items the grader coded as abstentions and the rule as answers. G = the grader's decisions scored with OR@0 correctness (\(R^{\mathrm{dec}} = t + (1-t)G\)). \(U - D = (G - D) + (U - G)\): \(G - D\) is the coding part and \(U - G\) the content part. A diagnostic fixed in advance; descriptive.}\label{A7_u_minus_d}
\input{tables/A7_u_minus_d.tex}
\end{table}
\begin{table}[htbp]
\centering\small
\caption{\textbf{Intervention study: predictions and plan items fixed in advance, their applicability and the observed values.} Predictions (1)--(3) were keyed to OLMo 3 Final's bands in the main study; its \(E_{\text{DiD}}\) band was ``in between'', so none applied, and none is judged right or wrong. The observed 2×2 differences (I1 seed 0, 95\% intervals) are descriptive. \(\mathrm{Gap}_k\) = selection regret against P(IK). The recovery ratio is \(U\)(TCAT, mean of three seeds)/\(U\)(I0) with a 95\% Fieller interval (TCAT seed variance/3 added), reported where \(U\)(I0)'s one-sided lower bound is above 0; the plan's appendix named I0-k as the regret reference instead (Table A19). The Qwen rows are one seed on 500 items, point estimates.}\label{A8_fix_predictions}
\input{tables/A8_fix_predictions.tex}
\end{table}
\begin{table}[htbp]
\centering\small
\caption{\textbf{SimpleQA Verified: the four final checkpoints beside the frontier models of \citet{kalai2026evaluating}.} Our rows: the first 200 SimpleQA Verified items under Kalai et al.'s protocol, at most 256 tokens, primary grader, 95\% paired intervals from 2,000 resamples; \(U\) at each condition's own \(t\) (0 = always-abstain). The last four rows are quoted from Kalai et al.'s Extended Data Table 3 (all 4,326 SimpleQA items, another grader): side by side, not a replication. All 24 OR@.75 answers of OLMo 3 Final were wrong. Its bridge result was partly known beforehand, because the intervention study reuses these replies.}\label{A9_simpleqa_bridge}
\input{tables/A9_simpleqa_bridge.tex}
\end{table}
\begin{table}[htbp]
\centering\small
\caption{\textbf{Sampled decisions (OR@.75, temperature 1.0) against greedy decoding.} Four final checkpoints, 100 items (50 + 50), five samples per item; greedy = the main run's OR@.75 reply. Differences are sampled − greedy, resampling each item's five samples together (95\% intervals, 2,000 resamples). U per sample = the range over the five sample indices; unanimous = all five samples made the same decision; greedy = majority = the greedy decision matches the majority of the five.}\label{A10_sampled_decisions}
\input{tables/A10_sampled_decisions.tex}
\end{table}
\begin{table}[htbp]
\centering\small
\caption{\textbf{Reasoning models: OLMo 3 Think and RL-Zero General.} Think: 100 items, at most 4,096 tokens; RL-Zero General: 200 items, at most 2,048 tokens. Instruct columns: OLMo 3 Instruct on the same items (main run, no intervals). Base: OLMo 3 Base's few-shot raw answers to the same items. Triplets are C / OR@.75 / OR@.9. Unfinished replies hit the token limit before closing the thinking block and are coded as abstentions; counted as wrong, \(U\)(.75)/\(U\)(.9) would be −0.94/−3.96 for Think and −0.77/−3.20 for RL-Zero. RL-Zero never closes a thinking block, so its whole transcript is graded; the primary grader coded 20\% of its OR@.75 replies as abstentions and the deterministic rule 1\%, a difference we did not resolve. Token signal = mean log-probability of the graded answer's tokens (only its AUROC is read).}\label{A11_reasoning_arm}
\input{tables/A11_reasoning_arm.tex}
\end{table}
\begin{table}[htbp]
\centering\small
\caption{\textbf{SmolLM3's mid-training checkpoint against its SFT checkpoint.} Mid did not follow the protocol's format: most of its replies under C and the rules hit the 256-token limit, only some belief answers were graded as attempts, its stated confidence was almost never parsed (so \(\Delta\) is undefined), and P(True) and P(IK) put less than half of the probability mass on the option letters for every item. Its positive \(U\)(OR@.75) comes from truncated replies graded as abstentions, not from compliance: abstention was flat in \(t\) and the probe response was 0. The SFT − Mid contrasts mostly reflect SFT teaching the format, so we do not read them as changes in calibration. \(\mathrm{Gap}_p\) = selection regret against P(True).}\label{A12_smollm3_mid}
\input{tables/A12_smollm3_mid.tex}
\end{table}
\begin{table}[htbp]
\centering\small
\caption{\textbf{Probe: effective thresholds \(c^*_t\), slopes and deference per primary checkpoint.} 50 probe items; the candidate answer is the model's own belief answer. At each \(t\), a logistic fit of \(P_{\text{give}}\) on logit \(c\) gives \(c^*_t\), where \(P_{\text{give}}\) = 1/2, with a 95\% Fieller interval \citep{fieller1954some}; following the rule means \(c^*_t = t\). † = outside the stated probabilities (0.20--0.95), extrapolated; values are clipped to {[}0, 1{]}. Slopes are on logit \(c\); near 0 (OLMo 2), \(c^*\) cannot be interpreted. A response with correctly placed thresholds and slope β scores \(E_{\text{DiD}}\) = 0.05, 0.22, 0.55 and 0.89 for β = 0.5, 1, 2 and 5. Deference share = \(\beta_c/(\beta_c + \beta_{\mathrm{own}})\) from a joint regression of \(P_{\text{give}}\) on (logit \(c\) − logit \(t\)) and (the model's own logit P(True) for the candidate − logit \(t\)): 1 means decisions follow the stated \(c\) only, 0.5 that \(c\) and the model's own P(True) weigh equally. Averaging the two option orders removes a position bias from the mean, but a model that partly favoured one position would show a smaller \(E_{\text{DiD}}\) and slope with its \(c^*_t\) unchanged.}\label{A13_probe_thresholds}
\input{tables/A13_probe_thresholds.tex}
\end{table}
\FloatBarrier
\subsection{Pre-specified analyses added after an earlier draft}
\begin{table}[htbp]
\centering\small
\caption{\textbf{Per checkpoint: the probe in context and in the abstract, H3 under both graders, and item-level monotonicity.} Point estimates (intervals for the 12-checkpoint means are in Table 1). Abstract probe: the same \(c \times t\) grid with no real question, one prompt per cell. \(\Delta\) (H3) is identified where it is defined, that is, where both stated-confidence groups have at least 20 items at some \(t\); Tulu 3 SFT and Final are not. Determ. columns: \(\Delta\) and OR@.75 abstention under the deterministic rule, which coded almost none of the OLMo 2, Tulu 3 and SmolLM3 replies as abstentions (primary-grader abstention in Table A2), so their \(\Delta\) is near 0 and the rule's mean is set by OLMo 3. Non-monotone = share of items whose decisions are not monotone in \(t\) (verbatim: \(t\) = .5, .75, .9, .95, .99; ABST: .5, .75, .9), which a deterministic policy acting on fixed beliefs cannot produce; replies cut off at the token limit count as graded. Under the first-sentence rule (the deterministic grader's analysis), the verbatim shares were 32.2, 40.9 and 53.7\% for the three OLMo 3 checkpoints and at most 2.3\% for the others, whose abstentions are mostly undeclared (Table A23).}\label{A14_probe_monotonicity}
\input{tables/A14_probe_monotonicity.tex}
\end{table}
\begin{table}[htbp]
\centering\small
\caption{\textbf{Selection regret with the drift checks fixed in advance, and under the deterministic grader.} Point estimates. \(\mathrm{Gap}_p\) and \(\mathrm{Gap}_k\) = selection regret against P(True) and P(IK) (Section 3.5). Same answer = restricted to items whose answer under C matches the belief answer; realized = answered items scored by their own OR@\(t\) correctness. Agree = share of OR@.75 answers identical to the OR@0 answer as strings, near 0 for verbose models, so the 90\% flag fixed in advance fires for every checkpoint and carries little information. The deterministic columns are defined for the four checkpoints that meet the inclusion rule (at least 20 abstentions and 20 answers at some \(t\)); on the same four, the primary grader's means are \(\mathrm{Gap}_p\) +0.097 and \(\mathrm{Gap}_k\) +0.002. Tulu 3 Final's deterministic values come from \(t\) = .5 alone. The deterministic rule also changes correctness grading (alias match).}\label{A15_gap_drift}
\input{tables/A15_gap_drift.tex}
\end{table}
\begin{table}[htbp]
\centering\small
\caption{\textbf{Truncation under OR@.75 and its bounds, and multiple-choice letter mass.} 1,000 items. Replies cut off at 256 tokens are graded as they are in the main analyses; the bounds grade all of them as abstentions or all as wrong. All three \(U\) values are below 0 for every checkpoint, so H1 does not depend on how truncation is treated. The other checks of H1 in Section 4.1, per checkpoint: without the 86 contamination-flagged items, \(U\) ranged from −1.13 to −0.24 (largest one-sided upper bound −0.19); under the deterministic grader, from −1.31 to −0.40 (−0.34); after the audit's difference correction (the grader's \(U\) plus the mean adjudicated-minus-grader score of the checkpoint's 20 audited replies), from −1.14 to −0.38 (largest upper 95\% bound −0.13). Last column: share of multiple-choice item rotations whose four option letters hold less than half of the probability mass. For the two-option read-outs the mean letter mass (over both orders) was at least 0.89 for P(True) and 0.92 for the probe in every checkpoint, with at most 0.4\% and 0\% of items below 0.5; for P(IK) it was lower in Tulu 3 DPO (0.39; 82\% of items below 0.5) and Tulu 3 SFT (0.62; 5\%). On items with mass of at least 0.5 (the design's check), the mean selection regret was +0.068 against P(True) and −0.046 against P(IK). H5 on the items whose letters held at least half of the mass in both the SFT and the PO checkpoint (the design's letter-mass restriction, applied to H5 post hoc): +0.093 {[}0.083, 0.103{]}, against +0.093 on all items; 486, 492, 125 and 308 of 500 items were kept for OLMo 3, OLMo 2, Tulu 3 and SmolLM3, with PO − SFT of +0.078, +0.136, +0.066 and +0.093, so Tulu 3's estimate rests on a quarter of its items. 8-bit control: on the 500 variant items, OLMo 3 7B Final run in 8 bits changed \(R\) at OR@.75 by −0.024 and abstention by −0.032 (the 32B models ran only in 8 bits).}\label{A16_truncation_mass}
\input{tables/A16_truncation_mass.tex}
\end{table}
\begin{table}[htbp]
\centering\small
\caption{\textbf{Context for H5: changes per recipe, temperature scaling and lineage estimates.} PO − SFT. Overconfidence = mean probability of the chosen option − accuracy. Temperature = the temperature fitted to the multiple-choice probabilities (higher means sharper original probabilities); ECE (expected calibration error) is measured on the original and on the temperature-scaled probabilities. On average the chosen option's probability rose by +0.088 {[}0.083, 0.092{]} and accuracy changed by −0.005 {[}−0.011, 0.002{]}. Calibration error after scaling and AUROC barely change, so the rise in overconfidence is consistent with a sharpening of the probabilities that leaves their ranking nearly unchanged. The band fixed in advance for H5, a ``material increase'' of at least 0.03 per recipe, was met in all four recipes. The last three rows estimate the change per preference-data lineage, as fixed in the design (OLMo 3's own preference data; Tulu 3's for OLMo 2, Tulu 3 and SmolLM3), and weighted by lineage (95\% intervals).}\label{A17_h5_context}
\input{tables/A17_h5_context.tex}
\end{table}
\begin{table}[htbp]
\centering\small
\caption{\textbf{Share of items TCAT answered under trained and held-out wordings (\%).} OLMo 3 7B; answered = not an abstention under the first-sentence rule. OR@.5 is a trained threshold and OR@.75 an unseen one, both in the trained wording. The other columns use wordings never used in training (500 items): BRIEF = OR@.75 with ``Keep your reply to a few words.'' before the question; PEN-L = ``Correct answers earn 1 point, wrong answers lose L points, and "I don't know" earns 0 points'' (implied \(t\) = L/(1 + L)); PARA = two paraphrases of the threshold; ABST = OR@.75 plus the ABST line.}\label{A18_tcat_wordings}
\input{tables/A18_tcat_wordings.tex}
\end{table}
\begin{table}[htbp]
\centering\small
\caption{\textbf{TCAT's training targets, validation answer shares and regret against two external references.} From stored outputs and training logs. \(\hat p > t\) = share of training items whose target is to answer (OLMo 3 and Qwen2.5); the threshold wording appeared in training only at \(t\) = .5 and .9 (OR@0 is the separate ``never abstain'' sentence). Regret = \(D\) of the reference policy − \(D\) of TCAT (mean of three seeds), split into a coverage part and a selection part; I0-k uses recalibrated P(IK), available before answering, and I0 recalibrated P(True) of the finished answer; the plan named I0-k as TCAT's regret reference. Validation = answered share / target answered share (\%) on up to 50 held-out validation items per trained condition at the end of training, for OLMo seeds 0, 1 and 2 and Qwen seed 0.}\label{A19_tcat_targets_regret}
\input{tables/A19_tcat_targets_regret.tex}
\end{table}
\begin{table}[htbp]
\centering\small
\caption{\textbf{I1's stated confidence after training: spread and validation error.} Counted from stored outputs and training logs. For REL's OR@0 answers (OLMo 1,000, Qwen 500): the number of distinct stated values each I1 adapter gave and the share of its most common value; and the mean absolute error against the targets on up to 50 held-out validation items at 25, 50, 75 and 100\% of training (0--100 scale). This is the collapse monitoring the plan required to be reported.}\label{A20_i1_spread}
\input{tables/A20_i1_spread.tex}
\end{table}
\begin{table}[htbp]
\centering\small
\caption{\textbf{F1--F7 per seed.} Pre-specified per-seed values and whether every seed's own bound passes. Notation as in Table 4: actual differences with one-sided 95\% bounds in parentheses (an upper bound for F6); a part passes when its bound is above −0.02 for F2(c), F4's AUROC part and F5, below +0.05 for F6, and above 0 otherwise. Pooled = mean of seeds (variance = bootstrap variance of the seed mean + seed variance/3); with 2 degrees of freedom for the seed variance, the normal approximation may be lenient. F4 uses three I1 seeds; F5's I1 parts use seed 0 only. F2 parts: (a) coverage strictly between the neighbouring thresholds (below, above), (b) closer agreement with I0 at that threshold than at the nearest trained one, (c) \(U\) ≥ −0.02.}\label{A21_fix_per_seed}
\input{tables/A21_fix_per_seed.tex}
\end{table}
\begin{table}[htbp]
\centering\small
\caption{\textbf{Decisions against the model's own stated confidence.} Point estimates. Each cell gives the \% of items where the OR@\(t\) decision agrees with comparing the stated confidence s (or \(s_E\), its expected value) with \(t\) / answered although s \textless{} t (over-answering) / declined although s \textgreater{} t (over-abstaining), over items whose belief answer was graded as an attempt and whose s is available. A tie s = t counts as agreement if the model abstained and in no column if it answered, so the shares fall short of 100 by the answered ties (SmolLM3 stated exactly 75 for 32--75\% of items). The high agreement of Tulu 3 and OLMo 2 reflects saturation: s was almost always above \(t\) and the models almost always answered.}\label{A22_own_confidence}
\input{tables/A22_own_confidence.tex}
\end{table}
\FloatBarrier
\subsection{Further post-hoc analyses}
\begin{table}[htbp]
\centering\small
\caption{\textbf{What the primary grader's abstentions consist of, and grader agreement by dimension.} Post-hoc analysis (f), specified and hashed before it ran; descriptive. Left: shares of replies (\%) under OR@.75 (1,000 items) and ABST@.75 (500 items). decl. \& graded = the first sentence declares an abstention (the deterministic rule's phrase list) and the primary grader judges NOT\_ATTEMPTED; graded only = judged NOT\_ATTEMPTED without a first-sentence declaration (in parentheses, the share of these cut off at the token limit); decl. only = declared but judged an attempt; discord. = the two discordant cells under ABST. Right: agreement (\%) with the primary grader on the stored second-grader sample (3,000 stratified replies and the 297 audit replies not among them, about 275 per checkpoint), on abstention (NOT\_ATTEMPTED or not) and on correctness among replies both graders treat as attempts; pooled κ: gpt-oss-20b 0.944 and 0.964, the rule 0.860 and 0.858.}\label{A23_abstention_coding}
\input{tables/A23_abstention_coding.tex}
\end{table}
\begin{table}[htbp]
\centering\small
\caption{\textbf{The external threshold calibrated on one dataset and tested on the other.} Post-hoc analysis (g), specified and hashed before it ran; means of the 12 primary checkpoints, primary grader, end to end. The isotonic map of P(True) is fitted on every item of the source dataset and applied to the other dataset's 250 variant items. Chosen = the format (ABST@t, verbatim OR@t or always-abstain) with the highest R on the source dataset's variant items, applied to the test items, so that both policies use the same labels. 95\% intervals from 2,000 resamples of source and test items, drawn separately within strata, refitting the map and re-choosing the format each time; in parentheses, the checkpoints whose interval is above / below 0. The wording fixed before the analysis: above 0 ``I0 higher'', below 0 ``lower'', otherwise ``no clear difference'' (not equivalence). Calibrated on TriviaQA, the map answered too often on PopQA; calibrated on PopQA, it was conservative on TriviaQA. The map is monotone, so the threshold acts as a cut on P(True), and transfer changes only where the cut falls, not how items are ranked. The two datasets differ in topic, difficulty and popularity at once, which this analysis does not separate.}\label{A24_transfer_calibration}
\input{tables/A24_transfer_calibration.tex}
\end{table}
\begin{table}[htbp]
\centering\small
\caption{\textbf{TCAT-dense: the threshold-conditioned decision trained with the threshold wording at four values.} Post-hoc analysis (h), specified and hashed before it ran; OLMo 3 7B, one seed, so there is no seed variance and no test. Trained as TCAT (same items, labels, \(\hat p\), settings and validation split) under C, OR@0 and OR@.3, .5, .7 and .9, 1,707 steps on 13,656 examples (TCAT: 1,138 on 9,104), so the number of thresholds and the amount of training change together; evaluated on the 1,000 items. Coverage = share answered under the first-sentence rule (\%); TCAT = mean of three seeds (not run at .4, .7 and .8); I0 = share whose recalibrated P(True) exceeds \(t\), the rule the training targets follow. Lower rows: the three readings fixed before the run at each unseen threshold, (a) coverage below the lower neighbour's and above the upper neighbour's (both above 0), (b) agreement with I0 at \(t\) minus the larger agreement with I0 at the neighbouring trained thresholds (above 0) and (c) \(U\) at least −0.02; TCAT-dense ``follows the unseen thresholds'' only if all three hold at all four, which they did not: (a) held at all four, (b) at none and (c) at all but .75. With trained thresholds closer together, (b) is stricter than F2's (b) was for TCAT: at .6 it must exceed the agreement with I0 at both .5 and .7, where F2's (b) compared only the nearest trained threshold, .5. Decisions not monotone over the nine thresholds: 0.1\% of items (TCAT: 14.6\% over six, mean of seeds). Regret in \(D\) (reference − TCAT-dense) at .5, .6 and .75: I0-k +0.006, −0.007 and +0.018; I0 +0.054, +0.040 and +0.037. Answered on the 500 variant items (\%): OR@.75 5.6, BRIEF@.75 93.0, PEN1 39.4, PEN3 39.8, PEN4 43.2. OR@0 accuracy minus REL's +0.008; abstention under C minus REL's −0.079. Coverage was below I0's at every trained threshold too (.3, .5, .7 and .9), where its training targets follow I0's rule, so TCAT-dense answered more conservatively than its targets.}\label{A25_tcat_dense}
\input{tables/A25_tcat_dense.tex}
\end{table}
\end{document}

%% file: math_commands.tex
\usepackage{amsmath,amsfonts,bm}

\def\eqref#1{equation~\ref{#1}}

\def\1{\bm{1}}

\DeclareMathAlphabet{\mathsfit}{\encodingdefault}{\sfdefault}{m}{sl}
\SetMathAlphabet{\mathsfit}{bold}{\encodingdefault}{\sfdefault}{bx}{n}



%% file: tables/T1_confirmatory.tex
\footnotesize%
\setlength{\tabcolsep}{4pt}%
\begin{tabularx}{\textwidth}{l>{\raggedright\arraybackslash}X>{\raggedleft\arraybackslash}p{1.7cm}rr>{\raggedright\arraybackslash}p{1.9cm}>{\raggedright\arraybackslash}p{3.1cm}}
\toprule
 & Hypothesis & Estimate & 95\% CI & \makecell{p \\{} (unadjusted)} & \makecell{effect $>$ 0 \\{} detected} & \makecell{band: point rule / \\{} interval rule} \\
\midrule
H1 & U(OR@.75) $<$ 0 for all 12 & 12/12 (U $-$1.08 to $-$0.23) & -- & $<$0.001 & yes (Holm 0.0083) & -- \\
H2 & threshold responsiveness $E_{\mathrm{DiD}}$ $>$ 0 & 0.037 & [0.034, 0.039] & $<$0.001 & yes (Holm 0.0100) & clear no / clear no \\
H3 & consistency with stated confidence $\Delta$ $>$ 0 (10 identified) & 0.062 & [0.029, 0.095] & 0.001 & yes (Holm 0.0167) & clear no / clear no \\
H4a & selection regret vs P(True) $>$ 0 & 0.068 & [0.012, 0.122] & 0.008 & yes (Holm 0.0250) & clear yes / in between \\
H4b & selection regret vs P(IK) $>$ 0 & $-$0.059 & [$-$0.119, 0.005] & 0.968 & no (Holm 0.0500) & clear no / clear no \\
H5 & MC overconfidence PO $-$ SFT $>$ 0 & 0.093 & [0.084, 0.101] & $<$0.001 & yes (Holm 0.0125) & -- \\
\midrule
H6 & abstain(ABST@.9) $-$ abstain(ABST@.5) $>$ 0 & $-$0.008 & [$-$0.012, $-$0.004] & 1.000 & separate family: not supported & clear no / clear no \\
\bottomrule
\end{tabularx}%

%% file: tables/T2_format_stakes.tex
\footnotesize%
\setlength{\tabcolsep}{4pt}%
\begin{tabular}{lrrrrrrrr}
\toprule
Model & \makecell{verb. \\{} none} & \makecell{verb. \\{} .5} & \makecell{verb. \\{} .75} & \makecell{verb. \\{} .9} & \makecell{ABST \\{} none} & \makecell{ABST \\{} .5} & \makecell{ABST \\{} .75} & \makecell{ABST \\{} .9} \\
\midrule
OLMo 3 Final & 12 & 80 & 70 & 68 & 58 & 97 & 97 & 98 \\
Tulu 3 Final & 2 & 8 & 4 & 3 & 23 & 82 & 76 & 78 \\
OLMo 2 Final & 1 & 5 & 6 & 5 & 81 & 95 & 94 & 94 \\
SmolLM3 Final & 5 & 12 & 10 & 8 & 48 & 92 & 93 & 94 \\
Mean of 12 primary & 4 & 24 & 20 & 19 & 54 & 93 & 91 & 92 \\
\midrule
OLMo 3.1 32B (8-bit) & 9 & 66 & 67 & 66 & 57 & 79 & 80 & 80 \\
OLMo 2 32B (8-bit) & 6 & 40 & 36 & 37 & 90 & 97 & 98 & 98 \\
Qwen3-8B & 7 & 8 & 6 & 5 & 52 & 90 & 91 & 89 \\
Qwen3-14B & 7 & 19 & 17 & 19 & 42 & 63 & 66 & 66 \\
Qwen2.5-7B & 10 & 53 & 54 & 59 & 53 & 88 & 89 & 92 \\
Mistral-7B v0.3 & 10 & 28 & 32 & 29 & 30 & 50 & 49 & 50 \\
Phi-4 (14B) & 6 & 17 & 16 & 14 & 18 & 38 & 40 & 41 \\
Granite-3.3-8B & 6 & 23 & 25 & 24 & 32 & 39 & 41 & 39 \\
\bottomrule
\end{tabular}%

%% file: tables/T3_external_vs_abst.tex
\footnotesize%
\setlength{\tabcolsep}{4pt}%
\begin{tabularx}{\textwidth}{lrrrrrr>{\raggedright\arraybackslash}X}
\toprule
t & \makecell{coverage \\{} ABST / I0} & \makecell{accuracy \\{} ABST / I0} & \makecell{$R_t - t$ \\{} ABST / I0} & \makecell{$R_t - t$ \\{} perfect \\{} selection} & \makecell{$\Delta$R = R(I0) $-$ R(ABST) \\{} [95\% CI]} & \makecell{checkpoints: \\{} interval \\{} $>$ 0 / $<$ 0} & reading \\
\midrule
0.5 & 0.07 / 0.46 & 0.82 / 0.71 & +0.025 / +0.092 & +0.237 & \makecell{+0.067 \\{} [+0.049, +0.082]} & 12 / 0 & I0 higher \\
0.75 & 0.09 / 0.17 & 0.84 / 0.82 & +0.008 / +0.012 & +0.119 & \makecell{+0.004 \\{} [$-$0.004, +0.010]} & 3 / 1 & uncertain: the interval includes 0 (not equivalence) \\
0.9 & 0.08 / 0.04 & 0.81 / 0.89 & $-$0.005 / +0.001 & +0.047 & \makecell{+0.006 \\{} [+0.001, +0.011]} & 3 / 0 & I0 higher (lower bound 0.001); both within 0.01 of always-abstain in R \\
\bottomrule
\end{tabularx}%

%% file: tables/T4_fix_hypotheses.tex
\footnotesize%
\setlength{\tabcolsep}{4pt}%
\begin{tabularx}{\textwidth}{l>{\raggedright\arraybackslash}p{3.5cm}>{\raggedright\arraybackslash}p{1.6cm}>{\raggedright\arraybackslash}Xr>{\raggedright\arraybackslash}p{1.5cm}}
\toprule
 & Hypothesis & Seeds & Key parts: value (one-sided 95\% bound) & p & \makecell{Supported \\{} (Holm, \\{} F1--F7)} \\
\midrule
F1 & TCAT beats always-abstain at unseen .75 (U and D) & TCAT $\times$3 & U: +0.004 (lower $-$0.003); D: +0.003 (lower $-$0.005) & 0.272 & no \\
F2 & TCAT reads unseen thresholds & TCAT $\times$3 & .3 (b): $-$0.243 (lower $-$0.269); .6 (b): +0.091 (lower +0.076); .75 (b): $-$0.077 (lower $-$0.092) & 1.000 & no \\
F3 & TCAT selects beyond a surface policy & TCAT $\times$3 & vs surface policy: +0.027 (lower +0.017) & $<$0.001 & yes \\
F4 & I1 calibrates and still ranks & I1 $\times$3 & bias gain: +0.499 (lower +0.442); Brier gain: +0.296 (lower +0.266); AUROC(I1) $-$ AUROC($\hat{p}$): $-$0.290 (lower $-$0.321) & 1.000 & no \\
F5 & no knowledge cost & TCAT $\times$3; I1 seed 0 & TCAT OR@0 acc.: +0.015 (lower +0.001); TCAT MC acc.: +0.002 (lower $-$0.004); I1 OR@0 acc.: $-$0.012 (lower $-$0.027); I1 MC acc.: $-$0.006 (lower $-$0.015) & 0.190 & no \\
F6 & no rise in abstention under C and P & TCAT $\times$3 & C: $-$0.066 (upper $-$0.051); P: $-$0.016 (upper $-$0.004) & $<$0.001 & yes \\
F7 & TCAT beats the best prompt-only arm at .75 & TCAT $\times$3 & vs best prompt: $-$0.002 (lower $-$0.022) & 0.564 & no \\
\bottomrule
\end{tabularx}%

%% file: tables/A1_grader_audit.tex
\footnotesize%
\setlength{\tabcolsep}{4pt}%
\begin{tabular}{lrrrr}
\toprule
Sample & primary (Qwen3-30B) & gpt-oss-20b & rule & Claude (blind) \\
\midrule
random OR@.75 sample (n=240) & 97.1 (0.95) & 95.4 (0.93) & 83.8 (0.74) & 99.2 (0.99) \\
grader disagreements (n=60) & 86.7 (0.79) & 83.3 (0.75) & 10.0 ($-$0.23) & 95.0 (0.92) \\
SFT stage (n=100) & 97.0 (0.95) & 94.0 (0.91) & 71.0 (0.55) & 98.0 (0.97) \\
PO stage (n=100) & 94.0 (0.91) & 91.0 (0.86) & 66.0 (0.47) & 97.0 (0.95) \\
Final stage (n=100) & 94.0 (0.91) & 94.0 (0.91) & 70.0 (0.54) & 100.0 (1.00) \\
\bottomrule
\end{tabular}%

%% file: tables/A2_stages.tex
\footnotesize%
\setlength{\tabcolsep}{4pt}%
\begin{tabular}{lrrrrrrr}
\toprule
Checkpoint & U(OR@.75) & \makecell{abstain \\{} .75 (\%)} & \makecell{MC \\{} overconf.} & s$\geq$.9 & \makecell{AUROC \\{} P(True)} & \makecell{AUROC \\{} P(IK)} & \makecell{AUROC \\{} $s_E$} \\
\midrule
OLMo 3 SFT & $-$0.79 & 48 & 0.184 & 0.98 & 0.74 & 0.69 & 0.57 \\
OLMo 3 DPO & $-$0.28 & 70 & 0.262 & 0.97 & 0.76 & 0.71 & 0.76 \\
OLMo 3 Final & $-$0.23 & 70 & 0.271 & 0.93 & 0.81 & 0.71 & 0.79 \\
Tulu 3 SFT & $-$0.58 & 5 & 0.189 & 0.99 & 0.70 & 0.75 & 0.67 \\
Tulu 3 DPO & $-$0.42 & 6 & 0.252 & 0.98 & 0.74 & 0.73 & 0.72 \\
Tulu 3 Final & $-$0.48 & 5 & 0.274 & 0.99 & 0.75 & 0.68 & 0.71 \\
OLMo 2 SFT & $-$0.92 & 3 & 0.103 & 1.00 & 0.70 & 0.72 & 0.61 \\
OLMo 2 DPO & $-$0.80 & 3 & 0.238 & 0.97 & 0.68 & 0.77 & 0.66 \\
OLMo 2 Final & $-$0.80 & 5 & 0.271 & 0.84 & 0.66 & 0.74 & 0.51 \\
SmolLM3 SFT & $-$1.08 & 7 & 0.160 & 0.45 & 0.82 & 0.76 & 0.83 \\
SmolLM3 APO & $-$0.94 & 12 & 0.254 & 0.20 & 0.80 & 0.74 & 0.77 \\
SmolLM3 Final & $-$0.88 & 10 & 0.230 & 0.64 & 0.83 & 0.75 & 0.78 \\
\bottomrule
\end{tabular}%

%% file: tables/A3_r_bases.tex
\footnotesize%
\setlength{\tabcolsep}{4pt}%
\begin{tabular}{lrrrrr}
\toprule
Model & own e2e & own dec. & ABST e2e & ABST dec. & P(True) thr. \\
\midrule
OLMo 3 Final & 0.692 & 0.682 & 0.753 & 0.752 & 0.761 \\
Tulu 3 Final & 0.631 & 0.639 & 0.763 & 0.775 & 0.771 \\
OLMo 2 Final & 0.550 & 0.552 & 0.756 & 0.756 & 0.755 \\
SmolLM3 Final & 0.531 & 0.529 & 0.751 & 0.760 & 0.781 \\
\midrule
OLMo 3.1 32B (8-bit) & 0.734 & 0.777 & 0.785 & 0.784 & 0.757 \\
OLMo 2 32B (8-bit) & 0.833 & 0.831 & 0.752 & 0.754 & 0.826 \\
\midrule
Qwen3-8B & 0.511 & 0.509 & 0.767 & 0.769 & 0.783 \\
Qwen3-14B & 0.625 & 0.645 & 0.776 & 0.792 & 0.810 \\
Qwen2.5-7B & 0.737 & 0.765 & 0.758 & 0.768 & 0.774 \\
Mistral-7B v0.3 & 0.721 & 0.737 & 0.748 & 0.768 & 0.781 \\
Phi-4 (14B) & 0.719 & 0.723 & 0.765 & 0.811 & 0.803 \\
Granite-3.3-8B & 0.674 & 0.678 & 0.750 & 0.752 & 0.769 \\
\bottomrule
\end{tabular}%

%% file: tables/A4_posthoc_equal_coverage.tex
\footnotesize%
\setlength{\tabcolsep}{3.5pt}%
\begin{tabular}{lrrrrrrrr}
\toprule
t & m & base & \makecell{acc \\{} $S_{\mathrm{ABST}}$} & \makecell{acc \\{} $S_P$} & \makecell{$\Delta$acc \\{} [95\%]} & \makecell{$\Delta R_{\mathrm{sel}}$ \\{} [95\%]} & \makecell{end-to-end gap \\{} [95\%]} & \makecell{answer-format \\{} effect [95\%]} \\
\midrule
0.5 & 37 & 0.47 & 0.849 & 0.887 & \makecell{+0.038 \\{} [$-$0.047, +0.108]} & \makecell{$-$0.0002 \\{} [$-$0.004, +0.003]} & \makecell{+0.063 \\{} [$-$0.036, +0.159]} & \makecell{$-$0.025 \\{} [$-$0.089, +0.038]} \\
0.75 & 45 & 0.47 & 0.853 & 0.873 & \makecell{+0.020 \\{} [$-$0.059, +0.086]} & \makecell{$-$0.002 \\{} [$-$0.007, +0.002]} & \makecell{+0.037 \\{} [$-$0.053, +0.124]} & \makecell{$-$0.017 \\{} [$-$0.075, +0.044]} \\
0.9 & 41 & 0.47 & 0.847 & 0.869 & \makecell{+0.021 \\{} [$-$0.056, +0.100]} & \makecell{$-$0.002 \\{} [$-$0.006, +0.002]} & \makecell{+0.058 \\{} [$-$0.036, +0.163]} & \makecell{$-$0.037 \\{} [$-$0.090, +0.011]} \\
\bottomrule
\end{tabular}%

%% file: tables/A5_posthoc_per_checkpoint.tex
\footnotesize%
\setlength{\tabcolsep}{4pt}%
\begin{tabular}{lrrr}
\toprule
Checkpoint & t = .5 & t = .75 & t = .9 \\
\midrule
\multicolumn{4}{l}{\textit{A. $\Delta$R = R(I0) $-$ R(ABST), end to end}} \\
OLMo 3 SFT & +0.030 [+0.002, +0.049] & $-$0.007 [$-$0.018, +0.006] & +0.002 [$-$0.003, +0.008] \\
OLMo 3 DPO & +0.039 [+0.012, +0.060] & $-$0.004 [$-$0.019, +0.011] & $-$0.001 [$-$0.010, +0.006] \\
OLMo 3 Final & +0.047 [+0.018, +0.073] & +0.002 [$-$0.016, +0.016] & $-$0.002 [$-$0.012, +0.006] \\
\midrule
Tulu 3 SFT & +0.088 [+0.060, +0.123] & $-$0.010 [$-$0.032, +0.014] & $-$0.005 [$-$0.015, +0.001] \\
Tulu 3 DPO & +0.079 [+0.046, +0.110] & $-$0.021 [$-$0.052, $-$0.002] & +0.003 [$-$0.011, +0.019] \\
Tulu 3 Final & +0.088 [+0.050, +0.118] & $-$0.001 [$-$0.026, +0.024] & +0.018 [+0.0004, +0.034] \\
\midrule
OLMo 2 SFT & +0.060 [+0.023, +0.083] & +0.008 [$-$0.006, +0.023] & +0.010 [+0.001, +0.021] \\
OLMo 2 DPO & +0.051 [+0.015, +0.081] & $-$0.001 [$-$0.019, +0.015] & +0.010 [$-$0.002, +0.022] \\
OLMo 2 Final & +0.028 [+0.002, +0.072] & $-$0.001 [$-$0.018, +0.007] & +0.002 [$-$0.007, +0.011] \\
\midrule
SmolLM3 SFT & +0.094 [+0.058, +0.116] & +0.025 [+0.0005, +0.039] & +0.013 [$-$0.003, +0.022] \\
SmolLM3 APO & +0.099 [+0.064, +0.123] & +0.021 [+0.003, +0.037] & +0.010 [$-$0.001, +0.016] \\
SmolLM3 Final & +0.099 [+0.075, +0.126] & +0.039 [+0.022, +0.053] & +0.017 [+0.006, +0.028] \\
\midrule
\multicolumn{4}{l}{\textit{B. $\Delta$acc = acc(P(True) top m) $-$ acc(ABST's m items), same answers}} \\
OLMo 3 SFT & +0.200 [$-$0.182, +0.562] & +0.182 [$-$0.214, +0.500] & +0.111 [$-$0.250, +0.500] \\
OLMo 3 DPO & +0.091 [$-$0.231, +0.417] & 0 [$-$0.312, +0.250] & +0.091 [$-$0.248, +0.400] \\
OLMo 3 Final & 0 [$-$0.429, +0.333] & 0 [$-$0.400, +0.273] & $-$0.083 [$-$0.500, +0.231] \\
\midrule
Tulu 3 SFT & $-$0.050 [$-$0.267, 0] & $-$0.158 [$-$0.281, $-$0.043] & $-$0.156 [$-$0.300, 0] \\
Tulu 3 DPO & $-$0.057 [$-$0.127, +0.028] & $-$0.095 [$-$0.155, $-$0.024] & $-$0.086 [$-$0.156, $-$0.018] \\
Tulu 3 Final & $-$0.080 [$-$0.160, +0.012] & $-$0.017 [$-$0.104, +0.052] & $-$0.028 [$-$0.122, +0.040] \\
\midrule
OLMo 2 SFT & +0.111 [$-$0.028, +0.212] & +0.098 [$-$0.023, +0.239] & +0.100 [$-$0.023, +0.250] \\
OLMo 2 DPO & $-$0.020 [$-$0.200, +0.106] & $-$0.038 [$-$0.200, +0.080] & $-$0.041 [$-$0.212, +0.082] \\
OLMo 2 Final & 0 [$-$0.188, +0.208] & 0 [$-$0.200, +0.167] & 0 [$-$0.194, +0.185] \\
\midrule
SmolLM3 SFT & +0.100 [$-$0.050, +0.333] & +0.147 [0, +0.429] & +0.174 [0, +0.389] \\
SmolLM3 APO & +0.115 [0, +0.250] & +0.091 [0, +0.227] & +0.143 [0, +0.375] \\
SmolLM3 Final & +0.050 [$-$0.034, +0.150] & +0.029 [$-$0.091, +0.154] & +0.032 [$-$0.088, +0.176] \\
\bottomrule
\end{tabular}%

%% file: tables/A6_models.tex
\footnotesize%
\setlength{\tabcolsep}{3.5pt}%
\begin{tabular}{lrrrrrrrr}
\toprule
Model & \makecell{U(OR@.75) \\{} [95\% CI]} & \makecell{ABST .9$-$.5 \\{} (pp)} & $E_{\mathrm{DiD}}$ & s$\geq$.9 & \makecell{AUROC \\{} P(True)} & \makecell{R(.75) \\{} own} & \makecell{R(.75) ABST \\{} (500 items)} & \makecell{R(.75) \\{} P(True) thr.} \\
\midrule
OLMo 3 Final & \makecell{$-$0.23 \\{} [$-$0.31, $-$0.16]} & +0.2 & 0.09 & 0.93 & 0.81 & 0.692 & 0.753 & 0.761 \\
Tulu 3 Final & \makecell{$-$0.48 \\{} [$-$0.58, $-$0.37]} & $-$4.2 & 0.05 & 0.99 & 0.75 & 0.631 & 0.763 & 0.771 \\
OLMo 2 Final & \makecell{$-$0.80 \\{} [$-$0.92, $-$0.69]} & $-$0.4 & 0.01 & 0.84 & 0.66 & 0.550 & 0.756 & 0.755 \\
SmolLM3 Final & \makecell{$-$0.88 \\{} [$-$0.99, $-$0.76]} & +1.8 & 0.03 & 0.64 & 0.83 & 0.531 & 0.751 & 0.781 \\
\midrule
OLMo 3.1 32B (8-bit) & \makecell{$-$0.07 \\{} [$-$0.16, 0.02]} & +0.8 & 0.13 & 0.82 & 0.74 & 0.734 & 0.785 & 0.757 \\
OLMo 2 32B (8-bit) & \makecell{0.33 \\{} [0.23, 0.42]} & +0.6 & 0.25 & 0.96 & 0.82 & 0.833 & 0.752 & 0.826 \\
\midrule
Qwen3-8B & \makecell{$-$0.96 \\{} [$-$1.12, $-$0.80]} & $-$0.6 & 0.15 & 0.91 & 0.84 & 0.511 & 0.767 & 0.783 \\
Qwen3-14B & \makecell{$-$0.50 \\{} [$-$0.64, $-$0.34]} & +3.4 & 0.46 & 0.87 & 0.88 & 0.625 & 0.776 & 0.810 \\
Qwen2.5-7B & \makecell{$-$0.05 \\{} [$-$0.16, 0.05]} & +4.8 & 0.29 & 0.94 & 0.75 & 0.737 & 0.758 & 0.774 \\
Mistral-7B v0.3 & \makecell{$-$0.12 \\{} [$-$0.24, 0.00]} & $-$0.4 & 0.08 & 0.97 & 0.76 & 0.721 & 0.748 & 0.781 \\
Phi-4 (14B) & \makecell{$-$0.12 \\{} [$-$0.27, 0.01]} & +2.6 & 0.31 & 0.95 & 0.85 & 0.719 & 0.765 & 0.803 \\
Granite-3.3-8B & \makecell{$-$0.30 \\{} [$-$0.44, $-$0.17]} & $-$0.4 & 0.10 & 0.93 & 0.77 & 0.674 & 0.750 & 0.769 \\
\bottomrule
\end{tabular}%

%% file: tables/A7_u_minus_d.tex
\footnotesize%
\setlength{\tabcolsep}{3pt}%
\begin{tabular}{lrrrrrrrrrrr}
\toprule
Arm & t & \makecell{both \\{} ans.} & \makecell{grader abst., \\{} rule ans.} & \makecell{grader ans., \\{} rule abst.} & \makecell{both \\{} abst.} & \makecell{OR@0 correct \\{} (col. 4)} & U & G & D & \makecell{G $-$ D \\{} (coding)} & \makecell{U $-$ G \\{} (content)} \\
\midrule
REL & .6 & 209 & 105 & 77 & 609 & 0.18 & $-$0.009 & $-$0.039 & $-$0.156 & +0.117 & +0.030 \\
REL & .75 & 234 & 143 & 64 & 559 & 0.18 & $-$0.234 & $-$0.270 & $-$0.543 & +0.273 & +0.036 \\
I1 (seed 0) & .6 & 57 & 79 & 13 & 851 & 0.32 & +0.008 & $-$0.005 & $-$0.061 & +0.056 & +0.012 \\
I1 (seed 0) & .75 & 56 & 83 & 23 & 838 & 0.31 & $-$0.021 & $-$0.041 & $-$0.189 & +0.148 & +0.020 \\
\bottomrule
\end{tabular}%

%% file: tables/A8_fix_predictions.tex
\footnotesize%
\setlength{\tabcolsep}{4pt}%
\begin{tabularx}{\textwidth}{>{\raggedright\arraybackslash}p{2.3cm}>{\raggedright\arraybackslash}p{2.5cm}>{\raggedright\arraybackslash}p{2.7cm}>{\raggedright\arraybackslash}p{1.9cm}>{\raggedright\arraybackslash}X}
\toprule
Item & Fixed in advance & Applicability check & Status & Observed (descriptive) \\
\midrule
Prediction (1): $E_{\mathrm{DiD}}$ `applies' & I1's own decisions move toward I1+t & OLMo 3 Final $E_{\mathrm{DiD}}$ 0.093 [0.078, 0.108]: band `in between'; needs point $\geq$ 0.30 & not applicable & -- \\
Prediction (2): $E_{\mathrm{DiD}}$ `does not apply' & I1 own $\approx$ REL $\ll$ I1+t & OLMo 3 Final $E_{\mathrm{DiD}}$ 0.093 [0.078, 0.108]: band `in between'; needs upper bound $<$ 0.10 & not applicable & description only: t = .6: I1 own $-$ REL: U +0.016 [$-$0.021, +0.055] (includes 0), D +0.094 [+0.058, +0.131] (above 0); t = .75: I1 own $-$ REL: U +0.213 [+0.147, +0.278] (above 0), D +0.354 [+0.285, +0.426] (above 0); t = .6: I1+t $-$ I1 own: U $-$0.009 [$-$0.029, +0.011] (includes 0), D +0.060 [+0.031, +0.090] (above 0); t = .75: I1+t $-$ I1 own: U +0.021 [$-$0.009, +0.054] (includes 0), D +0.189 [+0.136, +0.244] (above 0) \\
Prediction (3): $\mathrm{Gap}_k$ `material' & $\mathrm{Gap}_k$ after TCAT $<$ $\mathrm{Gap}_k$ of REL & OLMo 3 Final $\mathrm{Gap}_k$ 0.036; needs $\geq$ 0.05 & not applicable & -- \\
Rubric wording & used only if F2 holds & F2 & not used (F2 not supported) & -- \\
Self-knowledge wording & used only if both learners' own labels beat swapped labels & selection gap, lower bound $>$ 0 & not used; described as a learned difficulty prior & OLMo $-$0.0008 [$-$0.0028]; Qwen +0.0196 [+0.0106] \\
Recovery ratio U(TCAT)/U(I0) & reported only if U(I0)`s one-sided lower bound $>$ 0, with a Fieller interval & U(I0) bound & condition met at .5, .6 and .75 & t = 0.5: U(I0) lower +0.082, ratio 0.60 [0.34, 0.92]; t = 0.6: U(I0) lower +0.047, ratio 0.20 [0.04, 0.43]; t = 0.75: U(I0) lower +0.012, ratio 0.10 [$-$0.15, 0.68] \\
Qwen replication 2$\times$2 at OR@.75 & F1, F3 and the 2$\times$2 & -- & computed (points, as planned) & REL own: U $-$0.054, D $-$0.692; REL, raw s $>$ t: U $-$0.710, D $-$0.740; REL, recalibrated s $>$ t: U --, D 0.000; I1 own: U $-$0.128, D $-$0.948; I1+t: U +0.074, D +0.042 \\
\bottomrule
\end{tabularx}%

%% file: tables/A9_simpleqa_bridge.tex
\footnotesize%
\setlength{\tabcolsep}{3pt}%
\begin{tabular}{>{\raggedright\arraybackslash}p{2.6cm}rrrrrr}
\toprule
Model & \makecell{abstain C / .75 / .9 \\{} (\%)} & \makecell{U(.75) at OR@.75 \\{} [95\%]} & \makecell{U(.9) at OR@.9 \\{} [95\%]} & \makecell{abstain .9 $-$ .75 \\{} [95\%]} & \makecell{correct C \\{} (\%)} & \makecell{truncated C \\{} (\%)} \\
\midrule
OLMo 3 Final & 20 / 88 / 86 & \makecell{$-$0.36 \\{} [$-$0.51, $-$0.24]} & \makecell{$-$1.17 \\{} [$-$1.62, $-$0.76]} & \makecell{$-$0.015 \\{} [$-$0.055, +0.025]} & 6 & 27 \\
OLMo 2 Final & 2 / 11 / 12 & \makecell{$-$2.47 \\{} [$-$2.63, $-$2.29]} & \makecell{$-$7.33 \\{} [$-$7.80, $-$6.82]} & \makecell{+0.015 \\{} [$-$0.005, +0.035]} & 6 & 0 \\
Tulu 3 Final & 4 / 24 / 22 & \makecell{$-$1.95 \\{} [$-$2.15, $-$1.73]} & \makecell{$-$6.17 \\{} [$-$6.76, $-$5.58]} & \makecell{$-$0.030 \\{} [$-$0.055, $-$0.010]} & 11 & 1 \\
SmolLM3 Final & 6 / 18 / 20 & \makecell{$-$2.27 \\{} [$-$2.44, $-$2.07]} & \makecell{$-$6.65 \\{} [$-$7.19, $-$6.10]} & \makecell{+0.015 \\{} [$-$0.010, +0.040]} & 6 & 2 \\
\midrule
Gemini 3 Pro (Kalai et al.) & -- & $-$0.05 & $-$1.50 & -- & -- & -- \\
GPT-5 (Kalai et al.) & -- & $-$0.31 & $-$1.34 & -- & -- & -- \\
Grok 4 (Kalai et al.) & -- & $-$0.17 & $-$1.19 & -- & -- & -- \\
Claude Opus 4.5 (Kalai et al.) & -- & $-$0.20 & $-$1.01 & -- & -- & -- \\
\bottomrule
\end{tabular}%

%% file: tables/A10_sampled_decisions.tex
\footnotesize%
\setlength{\tabcolsep}{2.5pt}%
\begin{tabular}{>{\raggedright\arraybackslash}p{1.6cm}rrrrrrr}
\toprule
Model & \makecell{abstain (\%) \\{} greedy / \\{} sampled} & \makecell{$\Delta$ abstain \\{} [95\%]} & \makecell{U(.75) \\{} greedy / \\{} sampled} & \makecell{$\Delta$U \\{} [95\%]} & \makecell{U per \\{} sample \\{} (range)} & \makecell{unanimous \\{} (\%)} & \makecell{greedy = \\{} majority \\{} (\%)} \\
\midrule
OLMo 3 Final & 69 / 66 & \makecell{$-$0.028 \\{} [$-$0.088, +0.036]} & $-$0.29 / $-$0.40 & \makecell{$-$0.108 \\{} [$-$0.284, +0.074]} & $-$0.58 $\sim$ $-$0.30 & 45 & 82 \\
OLMo 2 Final & 4 / 5 & \makecell{+0.008 \\{} [$-$0.024, +0.040]} & $-$0.76 / $-$1.10 & \makecell{$-$0.336 \\{} [$-$0.586, $-$0.098]} & $-$1.21 $\sim$ $-$0.93 & 82 & 95 \\
Tulu 3 Final & 3 / 8 & \makecell{+0.046 \\{} [+0.018, +0.072]} & $-$0.55 / $-$0.59 & \makecell{$-$0.038 \\{} [$-$0.226, +0.140]} & $-$0.68 $\sim$ $-$0.37 & 79 & 97 \\
SmolLM3 Final & 12 / 16 & \makecell{+0.038 \\{} [$-$0.014, +0.088]} & $-$0.52 / $-$0.94 & \makecell{$-$0.422 \\{} [$-$0.652, $-$0.206]} & $-$1.10 $\sim$ $-$0.74 & 67 & 87 \\
\bottomrule
\end{tabular}%

%% file: tables/A11_reasoning_arm.tex
\footnotesize%
\setlength{\tabcolsep}{4pt}%
\begin{tabular}{>{\raggedright\arraybackslash}p{4.2cm}rrrr}
\toprule
Measure & \makecell{Think \\{} (Final)} & \makecell{Instruct \\{} (same items)} & \makecell{RL-Zero \\{} General} & \makecell{Instruct \\{} (same items)} \\
\midrule
items & 100 & 100 & 200 & 200 \\
unfinished, C / .75 / .9 (\%) & 18 / 17 / 22 & -- & 1 / 1 / 1 & -- \\
abstain, C / .75 / .9 (\%) & 21 / 35 / 38 & 11 / 69 / 73 & 18 / 20 / 16 & 14 / 70 / 72 \\
\makecell{U(.75) under OR@.75 \\{} [95\%]} & \makecell{$-$0.43 \\{} [$-$0.73, $-$0.15]} & $-$0.29 & \makecell{$-$0.73 \\{} [$-$0.96, $-$0.51]} & $-$0.12 \\
\makecell{U(.9) under OR@.9 \\{} [95\%]} & \makecell{$-$1.98 \\{} [$-$2.78, $-$1.24]} & $-$0.83 & \makecell{$-$3.11 \\{} [$-$3.74, $-$2.48]} & $-$0.62 \\
abstain(.9) $-$ abstain(.75) [95\%] & \makecell{+0.030 \\{} [$-$0.060, +0.120]} & -- & \makecell{$-$0.035 \\{} [$-$0.070, 0]} & -- \\
correct under C $-$ Base answers [95\%] & \makecell{$-$0.130 \\{} [$-$0.220, $-$0.030]} & -- & \makecell{$-$0.025 \\{} [$-$0.080, +0.030]} & -- \\
AUROC of the token signal, C answers [95\%] & \makecell{0.45 \\{} [0.33, 0.59]} & -- & \makecell{0.55 \\{} [0.46, 0.63]} & -- \\
AUROC of the Base's raw answers, same items & 0.89 & -- & 0.89 & -- \\
\bottomrule
\end{tabular}%

%% file: tables/A12_smollm3_mid.tex
\footnotesize%
\setlength{\tabcolsep}{4pt}%
\begin{tabular}{lrrr}
\toprule
Measure & Mid & SFT & SFT $-$ Mid [95\%] \\
\midrule
truncated, OR@.75 (\%) & 99 & 8 & -- \\
belief answers attempted (\%) & 29 & 96 & -- \\
stated confidence missing (\%) & 98 & 1 & -- \\
P(True) letter mass $<$ .5 (\%) & 100 & 0 & -- \\
abstain OR@.5 / OR@.9 (\%) & 72 / 73 & 10 / 6 & $-$0.047 [$-$0.074, $-$0.020] (of .9 $-$ .5) \\
U(OR@.75) & +0.158 [+0.118, +0.198] & $-$1.083 & $-$1.241 [$-$1.355, $-$1.139] \\
answered accuracy, OR@.75 & 0.90 & 0.46 & -- \\
abstain(P) & 0.668 & 0.380 & $-$0.288 [$-$0.339, $-$0.239] \\
mean P(True) $-$ accuracy & $-$0.423 & 0.278 & +0.701 [+0.665, +0.737] \\
$\mathrm{Gap}_p$ & $-$0.024 & 0.137 & +0.160 [+0.055, +0.273] \\
$E_{\mathrm{DiD}}$ (probe) & $-$0.001 & +0.022 & +0.023 [+0.018, +0.028] \\
MC native accuracy & 0.269 & 0.557 & -- \\
MC native overconfidence & 0.290 & 0.160 & $-$0.130 [$-$0.160, $-$0.098] \\
\bottomrule
\end{tabular}%

%% file: tables/A13_probe_thresholds.tex
\footnotesize%
\setlength{\tabcolsep}{3pt}%
\begin{tabular}{lrrrrrr}
\toprule
Checkpoint & c* no rule & c* t = .5 & c* t = .75 & c* t = .9 & \makecell{slope \\{} .5 / .75 / .9} & \makecell{deference \\{} share} \\
\midrule
OLMo 3 SFT & \makecell{0.20 \\{} [0.18, 0.23]} & \makecell{0.44 \\{} [0.40, 0.49]} & \makecell{0.51 \\{} [0.46, 0.56]} & \makecell{0.51 \\{} [0.47, 0.57]} & 0.66 / 0.61 / 0.61 & 0.98 \\
OLMo 3 DPO & \makecell{0.32 \\{} [0.28, 0.36]} & \makecell{0.52 \\{} [0.48, 0.57]} & \makecell{0.61 \\{} [0.56, 0.65]} & \makecell{0.65 \\{} [0.60, 0.70]} & 0.97 / 0.94 / 0.90 & 0.89 \\
OLMo 3 Final & \makecell{0.33 \\{} [0.29, 0.37]} & \makecell{0.55 \\{} [0.50, 0.61]} & \makecell{0.64 \\{} [0.59, 0.69]} & \makecell{0.68 \\{} [0.62, 0.73]} & 0.98 / 0.96 / 0.89 & 0.92 \\
\midrule
Tulu 3 SFT & \makecell{0.06$^\dagger$ \\{} [0.04, 0.08]} & \makecell{0.19$^\dagger$ \\{} [0.15, 0.24]} & \makecell{0.13$^\dagger$ \\{} [0.09, 0.17]} & \makecell{0.10$^\dagger$ \\{} [0.07, 0.15]} & 0.22 / 0.17 / 0.14 & 0.55 \\
Tulu 3 DPO & \makecell{0.02$^\dagger$ \\{} [0.01, 0.03]} & \makecell{0.14$^\dagger$ \\{} [0.11, 0.18]} & \makecell{0.08$^\dagger$ \\{} [0.06, 0.11]} & \makecell{0.06$^\dagger$ \\{} [0.04, 0.08]} & 0.43 / 0.35 / 0.29 & 0.52 \\
Tulu 3 Final & \makecell{0.01$^\dagger$ \\{} [0.01, 0.02]} & \makecell{0.11$^\dagger$ \\{} [0.09, 0.15]} & \makecell{0.07$^\dagger$ \\{} [0.05, 0.09]} & \makecell{0.05$^\dagger$ \\{} [0.03, 0.07]} & 0.49 / 0.39 / 0.33 & 0.52 \\
\midrule
OLMo 2 SFT & \makecell{0.67 \\{} [0.53, 0.79]} & \makecell{1.00$^\dagger$ \\{} [1.00, 1.00]} & \makecell{1.00$^\dagger$ \\{} [1.00, 1.00]} & \makecell{1.00$^\dagger$ \\{} [1.00, 1.00]} & 0.03 / 0.02 / 0.03 & 0.57 \\
OLMo 2 DPO & \makecell{0.10$^\dagger$ \\{} [0.06, 0.17]} & \makecell{0.38 \\{} [0.21, 0.66]} & \makecell{0.49 \\{} [0.30, 0.76]} & \makecell{0.69 \\{} [0.45, 0.91]} & 0.06 / 0.07 / 0.06 & 0.36 \\
OLMo 2 Final & \makecell{0.07$^\dagger$ \\{} [0.05, 0.11]} & \makecell{0.18$^\dagger$ \\{} [0.07, 0.42]} & \makecell{0.31 \\{} [0.14, 0.57]} & \makecell{0.46 \\{} [0.22, 0.74]} & 0.08 / 0.09 / 0.09 & 0.35 \\
\midrule
SmolLM3 SFT & \makecell{0.04$^\dagger$ \\{} [0.02, 0.07]} & \makecell{0.32 \\{} [0.22, 0.46]} & \makecell{0.52 \\{} [0.41, 0.65]} & \makecell{0.63 \\{} [0.50, 0.75]} & 0.23 / 0.26 / 0.23 & 0.42 \\
SmolLM3 APO & \makecell{0.25 \\{} [0.16, 0.38]} & \makecell{0.59 \\{} [0.42, 0.75]} & \makecell{0.66 \\{} [0.53, 0.79]} & \makecell{0.74 \\{} [0.60, 0.85]} & 0.30 / 0.34 / 0.30 & 0.44 \\
SmolLM3 Final & \makecell{0.20$^\dagger$ \\{} [0.13, 0.31]} & \makecell{0.40 \\{} [0.28, 0.54]} & \makecell{0.55 \\{} [0.44, 0.66]} & \makecell{0.62 \\{} [0.49, 0.74]} & 0.37 / 0.43 / 0.36 & 0.51 \\
\bottomrule
\end{tabular}%

%% file: tables/A14_probe_monotonicity.tex
\footnotesize%
\setlength{\tabcolsep}{3pt}%
\begin{tabular}{lrrrrrrrr}
\toprule
Checkpoint & \makecell{$E_{\mathrm{DiD}}$ \\{} contextual} & \makecell{$E_{\mathrm{DiD}}$ \\{} abstract} & \makecell{share \\{} s $\geq$ .9} & $\Delta$ (H3) & \makecell{$\Delta$ \\{} determ.} & \makecell{abstain \\{} OR@.75 \\{} determ. (\%)} & \makecell{non- \\{} monotone \\{} verbatim (\%)} & \makecell{non- \\{} monotone \\{} ABST (\%)} \\
\midrule
OLMo 3 SFT & 0.037 & 0.171 & 0.98 & 0.001 & $-$0.100 & 35.7 & 30.2 & 0.2 \\
OLMo 3 DPO & 0.085 & 0.139 & 0.97 & 0.008 & $-$0.080 & 72.5 & 36.8 & 0.8 \\
OLMo 3 Final & 0.093 & 0.155 & 0.93 & 0.130 & $-$0.187 & 62.3 & 40.1 & 0.8 \\
\midrule
Tulu 3 SFT & 0.030 & 0.300 & 0.99 & not identified & not identified & 0.3 & 8.5 & 3.6 \\
Tulu 3 DPO & 0.048 & 0.513 & 0.98 & 0.186 & 0.000 & 0.0 & 8.6 & 7.6 \\
Tulu 3 Final & 0.053 & 0.479 & 0.99 & not identified & $-$0.002 & 0.2 & 8.2 & 6.2 \\
\midrule
OLMo 2 SFT & 0.001 & 0.011 & 1.00 & $-$0.009 & 0.011 & 0.5 & 1.4 & 1.2 \\
OLMo 2 DPO & 0.005 & 0.029 & 0.97 & 0.038 & 0.000 & 0.2 & 2.4 & 1.2 \\
OLMo 2 Final & 0.006 & 0.043 & 0.84 & 0.010 & $-$0.001 & 0.3 & 2.5 & 0.8 \\
\midrule
SmolLM3 SFT & 0.022 & 0.058 & 0.45 & 0.103 & 0.007 & 0.4 & 8.3 & 1.6 \\
SmolLM3 APO & 0.027 & 0.103 & 0.20 & 0.075 & 0.000 & 0.2 & 8.0 & 0.2 \\
SmolLM3 Final & 0.035 & 0.177 & 0.64 & 0.079 & 0.000 & 0.1 & 7.6 & 0.2 \\
\midrule
Mean of 12 & 0.037 & 0.182 & -- & 0.062 (10) & $-$0.032 (11) & 14.4 & 13.5 & 2.0 \\
\bottomrule
\end{tabular}%

%% file: tables/A15_gap_drift.tex
\footnotesize%
\setlength{\tabcolsep}{2.5pt}%
\begin{tabular}{lrrrrrrrrr}
\toprule
Checkpoint & \makecell{agree \\{} OR@.75 \\{} (\%)} & $\mathrm{Gap}_p$ & \makecell{$\mathrm{Gap}_p$ \\{} same \\{} answer} & \makecell{$\mathrm{Gap}_p$ \\{} realized} & $\mathrm{Gap}_k$ & \makecell{$\mathrm{Gap}_k$ \\{} same \\{} answer} & \makecell{$\mathrm{Gap}_k$ \\{} realized} & \makecell{$\mathrm{Gap}_p$ \\{} determ.} & \makecell{$\mathrm{Gap}_k$ \\{} determ.} \\
\midrule
OLMo 3 SFT & 35.8 & +0.155 & +0.235 & +0.156 & +0.125 & +0.251 & +0.125 & +0.245 & +0.214 \\
OLMo 3 DPO & 0.7 & +0.071 & +0.017 & +0.072 & +0.042 & $-$0.006 & +0.042 & +0.107 & +0.095 \\
OLMo 3 Final & 0.0 & +0.076 & +0.083 & +0.050 & +0.036 & +0.064 & +0.010 & +0.136 & +0.118 \\
\midrule
Tulu 3 SFT & 65.8 & +0.178 & +0.165 & +0.267 & +0.164 & +0.268 & +0.253 & -- & -- \\
Tulu 3 DPO & 22.7 & +0.049 & $-$0.143 & +0.186 & $-$0.131 & +0.143 & +0.006 & -- & -- \\
Tulu 3 Final & 17.8 & +0.087 & +0.500 & +0.267 & $-$0.194 & +0.400 & $-$0.015 & +0.696 & +0.261 \\
\midrule
OLMo 2 SFT & 61.3 & +0.138 & $-$0.088 & +0.309 & $-$0.117 & $-$0.265 & +0.053 & -- & -- \\
OLMo 2 DPO & 34.6 & $-$0.290 & $-$0.571 & $-$0.355 & $-$0.226 & +0.143 & $-$0.290 & -- & -- \\
OLMo 2 Final & 38.0 & $-$0.084 & $-$0.185 & +0.065 & $-$0.194 & +0.037 & $-$0.045 & -- & -- \\
\midrule
SmolLM3 SFT & 23.5 & +0.137 & $-$0.100 & +0.207 & $-$0.047 & $-$0.333 & +0.023 & -- & -- \\
SmolLM3 APO & 11.7 & +0.188 & 0.000 & +0.372 & $-$0.009 & 0.000 & +0.175 & -- & -- \\
SmolLM3 Final & 17.9 & +0.107 & +0.250 & +0.183 & $-$0.160 & +0.250 & $-$0.084 & -- & -- \\
\midrule
Mean of 12 & -- & +0.068 & +0.013 & +0.148 & $-$0.059 & +0.079 & +0.021 & +0.296 & +0.172 \\
\bottomrule
\end{tabular}%

%% file: tables/A16_truncation_mass.tex
\footnotesize%
\setlength{\tabcolsep}{4pt}%
\begin{tabular}{lrrrrr}
\toprule
Checkpoint & \makecell{truncated \\{} (\%)} & \makecell{U(OR@.75) \\{} as graded} & \makecell{U, truncated \\{} as abstention} & \makecell{U, truncated \\{} as wrong} & \makecell{MC letter \\{} mass $<$ .5 (\%)} \\
\midrule
OLMo 3 SFT & 1.2 & $-$0.79 & $-$0.78 & $-$0.82 & 10.2 \\
OLMo 3 DPO & 5.5 & $-$0.28 & $-$0.21 & $-$0.37 & 25.2 \\
OLMo 3 Final & 11.2 & $-$0.23 & $-$0.14 & $-$0.48 & 21.6 \\
\midrule
Tulu 3 SFT & 0.1 & $-$0.58 & $-$0.58 & $-$0.58 & 15.8 \\
Tulu 3 DPO & 0.8 & $-$0.42 & $-$0.43 & $-$0.45 & 37.3 \\
Tulu 3 Final & 0.8 & $-$0.48 & $-$0.48 & $-$0.50 & 24.7 \\
\midrule
OLMo 2 SFT & 0.1 & $-$0.92 & $-$0.92 & $-$0.92 & 4.0 \\
OLMo 2 DPO & 0.1 & $-$0.80 & $-$0.79 & $-$0.80 & 3.6 \\
OLMo 2 Final & 0.1 & $-$0.80 & $-$0.80 & $-$0.80 & 3.0 \\
\midrule
SmolLM3 SFT & 8.2 & $-$1.08 & $-$0.98 & $-$1.23 & 25.5 \\
SmolLM3 APO & 3.3 & $-$0.94 & $-$0.89 & $-$0.99 & 35.0 \\
SmolLM3 Final & 1.4 & $-$0.88 & $-$0.85 & $-$0.90 & 21.4 \\
\bottomrule
\end{tabular}%

%% file: tables/A17_h5_context.tex
\footnotesize%
\setlength{\tabcolsep}{4pt}%
\begin{tabular}{lrrrrr}
\toprule
Recipe or lineage & \makecell{overconfidence \\{} PO $-$ SFT} & \makecell{temperature \\{} SFT $\rightarrow$ PO} & \makecell{ECE raw \\{} change} & \makecell{ECE after \\{} scaling, change} & \makecell{AUROC \\{} change} \\
\midrule
OLMo 3 & +0.078 & 2.19 $\rightarrow$ 3.57 & +0.084 & $-$0.006 & $-$0.015 \\
OLMo 2 & +0.135 & 1.49 $\rightarrow$ 2.66 & +0.129 & $-$0.004 & $-$0.008 \\
Tulu 3 & +0.064 & 2.13 $\rightarrow$ 3.12 & +0.062 & +0.006 & $-$0.016 \\
SmolLM3 & +0.094 & 1.95 $\rightarrow$ 3.14 & +0.086 & $-$0.015 & +0.007 \\
\midrule
OLMo 3 preference data & +0.078 [+0.065, +0.093] & -- & -- & -- & -- \\
Tulu 3 preference data & +0.097 [+0.088, +0.107] & -- & -- & -- & -- \\
lineage-weighted & +0.088 [+0.079, +0.098] & -- & -- & -- & -- \\
\bottomrule
\end{tabular}%

%% file: tables/A18_tcat_wordings.tex
\footnotesize%
\setlength{\tabcolsep}{3pt}%
\begin{tabular}{lrrrrrrrrr}
\toprule
Arm & \makecell{OR@.5 \\{} (trained)} & \makecell{OR@.75 \\{} (unseen)} & \makecell{BRIEF \\{} @.75} & \makecell{PEN-1 \\{} (t = .5)} & \makecell{PEN-3 \\{} (t = .75)} & \makecell{PEN-4 \\{} (t = .8)} & \makecell{PARA \\{} @.75A} & \makecell{PARA \\{} @.75B} & \makecell{ABST \\{} @.75} \\
\midrule
REL & 21.1 & 37.7 & 96.8 & 40.4 & 42.2 & 45.6 & 2.4 & 92.8 & 3.0 \\
TCAT seed 0 & 31.4 & 1.1 & 94.0 & 37.8 & 36.8 & 40.8 & 1.8 & 35.6 & 7.4 \\
TCAT seed 1 & 29.7 & 1.0 & 92.6 & 31.6 & 31.8 & 35.6 & 1.0 & 46.6 & 6.6 \\
TCAT seed 2 & 33.3 & 1.1 & 94.2 & 37.4 & 38.8 & 41.6 & 2.2 & 29.8 & 5.0 \\
\bottomrule
\end{tabular}%

%% file: tables/A19_tcat_targets_regret.tex
\footnotesize%
\setlength{\tabcolsep}{3pt}%
\begin{tabularx}{\textwidth}{>{\raggedright\arraybackslash}p{0.6cm}rrrr>{\raggedright\arraybackslash}X}
\toprule
t & \makecell{$\hat{p}$ $>$ t \\{} OLMo 3 (\%)} & \makecell{$\hat{p}$ $>$ t \\{} Qwen2.5 (\%)} & \makecell{regret vs I0-k \\{} (coverage / selection)} & \makecell{regret vs I0 \\{} (coverage / selection)} & validation answered / target (\%) \\
\midrule
0.3 & 79.2 & 86.4 & -- & -- & -- \\
0.5 & 55.9 & 68.5 & +0.003 (+0.000 / +0.003) & +0.051 ($-$0.003 / +0.054) & 40.0/48.0; 38.0/44.0; 44.0/60.0; 66.0/60.0 \\
0.6 & 43.9 & 61.6 & +0.022 (+0.011 / +0.010) & +0.068 (+0.054 / +0.013) & -- \\
0.75 & 23.3 & 50.0 & +0.014 (+0.012 / +0.002) & +0.033 (+0.035 / $-$0.002) & -- \\
0.9 & 3.3 & 29.0 & -- & -- & 0.0/2.0; 4.0/2.0; 0.0/2.0; 22.0/26.0 \\
\bottomrule
\end{tabularx}%

%% file: tables/A20_i1_spread.tex
\footnotesize%
\setlength{\tabcolsep}{4pt}%
\begin{tabular}{lrrrr}
\toprule
Adapter & answers & \makecell{distinct \\{} values} & \makecell{most common \\{} value (share)} & \makecell{validation MAE \\{} (25/50/75/100\%)} \\
\midrule
OLMo 3 I1 seed 0 & 1000 & 3 & 0.29 (99.7\%) & 27.6 $\rightarrow$ 17.7 $\rightarrow$ 16.1 $\rightarrow$ 28.5 \\
OLMo 3 I1 seed 1 & 1000 & 4 & 0.29 (98.9\%) & 24.5 $\rightarrow$ 23.5 $\rightarrow$ 20.4 $\rightarrow$ 26.7 \\
OLMo 3 I1 seed 2 & 1000 & 3 & 0.30 (62.8\%) & 25.3 $\rightarrow$ 19.6 $\rightarrow$ 24.4 $\rightarrow$ 20.2 \\
Qwen2.5 I1 seed 0 & 500 & 14 & 0.46 (29.0\%) & 10.4 $\rightarrow$ 10.7 $\rightarrow$ 8.7 $\rightarrow$ 7.2 \\
\bottomrule
\end{tabular}%

%% file: tables/A21_fix_per_seed.tex
\scriptsize%
\setlength{\tabcolsep}{3pt}%
\begin{tabularx}{\textwidth}{ll>{\raggedright\arraybackslash}p{3.0cm}>{\raggedright\arraybackslash}X>{\raggedright\arraybackslash}p{1.5cm}}
\toprule
 & part & pooled & per seed (0; 1; 2) & \makecell{every seed \\{} passes} \\
\midrule
F1 & U & +0.004 (lower $-$0.003) & +0.003 (lower $-$0.005); +0.006 (lower $-$0.001); +0.002 (lower $-$0.006) & no \\
F1 & D & +0.003 (lower $-$0.005) & +0.003 (lower $-$0.005); +0.006 (lower $-$0.001); $-$0.001 (lower $-$0.011) & no \\
F2 & .3 (a) below & +0.819 (lower +0.791) & +0.796 (lower +0.776); +0.843 (lower +0.825); +0.819 (lower +0.799) & yes \\
F2 & .3 (a) above & $-$0.145 (lower $-$0.169) & $-$0.124 (lower $-$0.141); $-$0.149 (lower $-$0.166); $-$0.163 (lower $-$0.181) & no \\
F2 & .3 (b) & $-$0.243 (lower $-$0.269) & $-$0.236 (lower $-$0.261); $-$0.250 (lower $-$0.276); $-$0.244 (lower $-$0.269) & no \\
F2 & .3 (c) & +0.093 (lower +0.075) & +0.101 (lower +0.083); +0.085 (lower +0.068); +0.093 (lower +0.076) & yes \\
F2 & .6 (a) below & +0.275 (lower +0.248) & +0.272 (lower +0.251); +0.257 (lower +0.236); +0.297 (lower +0.275) & yes \\
F2 & .6 (a) above & +0.029 (lower +0.021) & +0.031 (lower +0.022); +0.030 (lower +0.021); +0.025 (lower +0.017) & yes \\
F2 & .6 (b) & +0.091 (lower +0.076) & +0.092 (lower +0.077); +0.092 (lower +0.077); +0.088 (lower +0.073) & yes \\
F2 & .6 (c) & +0.015 (lower +0.005) & +0.015 (lower +0.002); +0.019 (lower +0.008); +0.013 (lower +0.001) & yes \\
F2 & .75 (a) below & +0.029 (lower +0.021) & +0.031 (lower +0.022); +0.030 (lower +0.021); +0.025 (lower +0.017) & yes \\
F2 & .75 (a) above & +0.011 (lower +0.006) & +0.011 (lower +0.006); +0.010 (lower +0.005); +0.011 (lower +0.006) & yes \\
F2 & .75 (b) & $-$0.077 (lower $-$0.092) & $-$0.076 (lower $-$0.091); $-$0.078 (lower $-$0.092); $-$0.078 (lower $-$0.092) & no \\
F2 & .75 (c) & +0.004 (lower $-$0.003) & +0.003 (lower $-$0.005); +0.006 (lower $-$0.001); +0.002 (lower $-$0.006) & yes \\
F3 & vs surface & +0.027 (lower +0.017) & +0.029 (lower +0.018); +0.028 (lower +0.018); +0.024 (lower +0.014) & yes \\
F4 & bias & +0.499 (lower +0.442) & +0.473 (lower +0.425); +0.474 (lower +0.426); +0.549 (lower +0.525) & yes \\
F4 & brier & +0.296 (lower +0.266) & +0.295 (lower +0.263); +0.296 (lower +0.264); +0.297 (lower +0.269) & yes \\
F4 & auroc & $-$0.290 (lower $-$0.321) & $-$0.303 (lower $-$0.327); $-$0.298 (lower $-$0.323); $-$0.270 (lower $-$0.305) & no \\
F5 & TCAT OR@0 & +0.015 (lower +0.001) & +0.022 (lower +0.008); +0.005 (lower $-$0.009); +0.019 (lower +0.005) & yes \\
F5 & TCAT MC & +0.002 (lower $-$0.004) & $-$0.002 (lower $-$0.007); +0.001 (lower $-$0.005); +0.005 (lower $-$0.001) & yes \\
F5 & I1 OR@0 & $-$0.012 (lower $-$0.027) & -- & -- \\
F5 & I1 MC & $-$0.006 (lower $-$0.015) & -- & -- \\
F6 & C & $-$0.066 (upper $-$0.051) & $-$0.065 (upper $-$0.049); $-$0.064 (upper $-$0.049); $-$0.068 (upper $-$0.052) & yes \\
F6 & P & $-$0.016 (upper $-$0.004) & $-$0.020 (upper $-$0.008); $-$0.010 (upper $-$0.000); $-$0.018 (upper $-$0.006) & yes \\
F7 & vs best prompt & $-$0.002 (lower $-$0.022) & $-$0.002 (lower $-$0.023); $-$0.002 (lower $-$0.023); $-$0.002 (lower $-$0.022) & no \\
\bottomrule
\end{tabularx}%

%% file: tables/A22_own_confidence.tex
\footnotesize%
\setlength{\tabcolsep}{4pt}%
\begin{tabular}{lrrrrr}
\toprule
Checkpoint & \makecell{s, t = .75 \\{} agree / over-ans. / \\{} over-abst.} & s, t = .9 & $s_E$, t = .75 & $s_E$, t = .9 & items \\
\midrule
OLMo 3 SFT & 53 / 1 / 46 & 61 / 1 / 38 & 54 / 1 / 45 & 58 / 33 / 9 & 972 \\
OLMo 3 DPO & 31 / 0 / 68 & 34 / 1 / 65 & 32 / 1 / 68 & 43 / 5 / 51 & 973 \\
OLMo 3 Final & 34 / 1 / 65 & 39 / 1 / 60 & 35 / 1 / 64 & 50 / 6 / 45 & 948 \\
\midrule
Tulu 3 SFT & 96 / 1 / 3 & 95 / 1 / 4 & 95 / 2 / 3 & 92 / 5 / 3 & 977 \\
Tulu 3 DPO & 96 / 1 / 3 & 95 / 1 / 4 & 96 / 1 / 3 & 94 / 2 / 4 & 964 \\
Tulu 3 Final & 96 / 1 / 3 & 96 / 1 / 3 & 96 / 1 / 3 & 96 / 1 / 3 & 981 \\
\midrule
OLMo 2 SFT & 98 / 0 / 2 & 93 / 0 / 2 & 95 / 3 / 2 & 71 / 27 / 2 & 987 \\
OLMo 2 DPO & 96 / 2 / 2 & 90 / 3 / 2 & 95 / 3 / 2 & 88 / 10 / 2 & 955 \\
OLMo 2 Final & 89 / 8 / 3 & 70 / 15 / 2 & 90 / 7 / 3 & 62 / 36 / 2 & 969 \\
\midrule
SmolLM3 SFT & 48 / 7 / 1 & 47 / 53 / 1 & 64 / 36 / 1 & 4 / 96 / 0 & 948 \\
SmolLM3 APO & 25 / 3 / 0 & 23 / 76 / 0 & 86 / 12 / 3 & 8 / 92 / 0 & 885 \\
SmolLM3 Final & 66 / 3 / 1 & 65 / 34 / 1 & 91 / 7 / 2 & 22 / 78 / 0 & 901 \\
\bottomrule
\end{tabular}%

%% file: tables/A23_abstention_coding.tex
\scriptsize%
\setlength{\tabcolsep}{2pt}%
\begin{tabular}{lrrrrrrrrr}
\toprule
Checkpoint & \makecell{OR@.75 \\{} decl. \& \\{} graded} & \makecell{OR@.75 \\{} graded only \\{} (trunc.)} & \makecell{OR@.75 \\{} decl. \\{} only} & \makecell{ABST@.75 \\{} decl. \& \\{} graded} & \makecell{ABST@.75 \\{} discord.} & \makecell{gpt-oss \\{} abst.} & \makecell{gpt-oss \\{} correct.} & \makecell{rule \\{} abst.} & \makecell{rule \\{} correct.} \\
\midrule
OLMo 3 SFT & 35.1 & 12.6 (3.2) & 0.6 & 97.8 & 0.2 & 98.2 & 98.3 & 92.0 & 93.2 \\
OLMo 3 DPO & 60.7 & 9.8 (10.2) & 11.8 & 96.8 & 0.4 & 96.7 & 99.0 & 88.7 & 93.8 \\
OLMo 3 Final & 55.9 & 14.3 (24.5) & 6.4 & 96.8 & 0.4 & 95.3 & 92.2 & 86.9 & 90.7 \\
\midrule
Tulu 3 SFT & 0.1 & 4.9 (2.0) & 0.2 & 92.4 & 0.2 & 98.2 & 99.5 & 95.6 & 90.4 \\
Tulu 3 DPO & 0.0 & 6.1 (3.3) & 0.0 & 70.0 & 0.6 & 97.5 & 97.2 & 95.6 & 93.1 \\
Tulu 3 Final & 0.1 & 5.1 (3.9) & 0.1 & 75.8 & 0.8 & 97.8 & 99.1 & 94.9 & 94.8 \\
\midrule
OLMo 2 SFT & 0.5 & 2.2 (4.5) & 0.0 & 91.8 & 0.2 & 99.6 & 99.1 & 98.9 & 94.3 \\
OLMo 2 DPO & 0.1 & 3.1 (0.0) & 0.1 & 89.4 & 1.0 & 98.9 & 99.1 & 97.1 & 93.8 \\
OLMo 2 Final & 0.2 & 4.5 (0.0) & 0.1 & 94.2 & 0.4 & 97.8 & 99.0 & 97.1 & 95.9 \\
\midrule
SmolLM3 SFT & 0.3 & 6.8 (13.2) & 0.1 & 96.6 & 0.2 & 96.7 & 97.3 & 93.1 & 84.7 \\
SmolLM3 APO & 0.0 & 12.5 (4.0) & 0.2 & 95.6 & 0.2 & 96.0 & 97.9 & 93.1 & 93.8 \\
SmolLM3 Final & 0.0 & 10.5 (1.0) & 0.1 & 92.8 & 0.4 & 96.7 & 97.9 & 92.4 & 95.2 \\
\midrule
Mean / pooled & 12.7 & 7.7 & 1.6 & 90.8 & 0.4 & 97.5 & 98.2 & 93.8 & 92.9 \\
\bottomrule
\end{tabular}%

%% file: tables/A24_transfer_calibration.tex
\scriptsize%
\setlength{\tabcolsep}{3pt}%
\begin{tabular}{lrrrrrrr}
\toprule
Calibrated on $\rightarrow$ tested on & t & \makecell{coverage \\{} I0} & \makecell{$R_t - t$ \\{} I0} & \makecell{$R_t - t$ \\{} ABST} & \makecell{$R_t - t$ \\{} chosen} & \makecell{$\Delta$R = I0 $-$ ABST [95\% CI] \\{} (checkpoints $>$ 0 / $<$ 0)} & \makecell{$\Delta$R = I0 $-$ chosen \\{} [95\% CI]} \\
\midrule
TriviaQA $\rightarrow$ PopQA & 0.5 & 0.53 & $-$0.011 & +0.013 & $-$0.062 & $-$0.024 [$-$0.061, +0.004] (1/3) & +0.051 [+0.029, +0.075] \\
TriviaQA $\rightarrow$ PopQA & 0.75 & 0.29 & $-$0.066 & +0.005 & +0.005 & $-$0.071 [$-$0.091, $-$0.033] (0/6) & $-$0.071 [$-$0.086, $-$0.002] \\
TriviaQA $\rightarrow$ PopQA & 0.9 & 0.03 & $-$0.002 & $-$0.002 & +0.0002 & $-$0.0003 [$-$0.023, +0.002] (0/1) & $-$0.002 [$-$0.024, +0.001] \\
\midrule
PopQA $\rightarrow$ TriviaQA & 0.5 & 0.45 & +0.142 & +0.038 & +0.034 & +0.104 [+0.076, +0.124] (9/0) & +0.107 [+0.078, +0.127] \\
PopQA $\rightarrow$ TriviaQA & 0.75 & 0.11 & +0.015 & +0.012 & +0.011 & +0.003 [$-$0.007, +0.015] (2/1) & +0.004 [$-$0.004, +0.016] \\
PopQA $\rightarrow$ TriviaQA & 0.9 & 0.06 & +0.002 & $-$0.009 & $-$0.005 & +0.011 [+0.0003, +0.020] (4/0) & +0.007 [$-$0.002, +0.014] \\
\bottomrule
\end{tabular}%

%% file: tables/A25_tcat_dense.tex
\scriptsize%
\setlength{\tabcolsep}{3pt}%
\begin{tabular}{lrrrrrr}
\toprule
t &  & \makecell{coverage \\{} TCAT-dense} & \makecell{coverage \\{} TCAT} & \makecell{coverage \\{} I0} & \makecell{U TCAT-dense \\{} [95\% CI]} & \makecell{D \\{} TCAT-dense} \\
\midrule
0.3 & trained & 63.8 & 16.9 & 67.8 & +0.144 [+0.107, +0.178] & +0.137 \\
0.4 & unseen & 53.7 & -- & 47.0 & +0.084 [+0.047, +0.122] & +0.075 \\
0.5 & trained & 27.2 & 31.5 & 36.4 & +0.065 [+0.032, +0.095] & +0.058 \\
0.6 & unseen & 18.9 & 3.9 & 26.6 & +0.048 [+0.017, +0.080] & +0.039 \\
0.7 & trained & 6.8 & -- & 18.2 & +0.021 [$-$0.002, +0.043] & +0.011 \\
0.75 & unseen & 4.7 & 1.1 & 10.0 & $-$0.001 [$-$0.026, +0.021] & $-$0.001 \\
0.8 & unseen & 1.6 & -- & 8.5 & +0.001 [$-$0.016, +0.014] & +0.001 \\
0.9 & trained & 0.1 & 0.0 & 1.4 & +0.001 [0, +0.003] & +0.001 \\
0.95 & unseen & 0.2 & 0.1 & 0.7 & +0.002 [0, +0.005] & +0.002 \\
\midrule
0.4 & readings & \makecell{(a) below +0.101 \\{} [+0.083, +0.120]} & \makecell{(a) above +0.265 \\{} [+0.238, +0.294]} & \makecell{(b) $-$0.006 \\{} [$-$0.029, +0.009]} & \makecell{(c) +0.084 \\{} [+0.047, +0.122]} & \makecell{a, b, c: \\{} yes, no, yes} \\
0.6 & readings & \makecell{(a) below +0.083 \\{} [+0.067, +0.100]} & \makecell{(a) above +0.121 \\{} [+0.102, +0.142]} & \makecell{(b) $-$0.038 \\{} [$-$0.057, $-$0.021]} & \makecell{(c) +0.048 \\{} [+0.017, +0.080]} & \makecell{a, b, c: \\{} yes, no, yes} \\
0.75 & readings & \makecell{(a) below +0.021 \\{} [+0.013, +0.030]} & \makecell{(a) above +0.031 \\{} [+0.021, +0.042]} & \makecell{(b) $-$0.050 \\{} [$-$0.069, $-$0.032]} & \makecell{(c) $-$0.001 \\{} [$-$0.026, +0.021]} & \makecell{a, b, c: \\{} yes, no, no} \\
0.8 & readings & \makecell{(a) below +0.031 \\{} [+0.021, +0.042]} & \makecell{(a) above +0.015 \\{} [+0.008, +0.023]} & \makecell{(b) $-$0.057 \\{} [$-$0.074, $-$0.042]} & \makecell{(c) +0.001 \\{} [$-$0.016, +0.014]} & \makecell{a, b, c: \\{} yes, no, yes} \\
\bottomrule
\end{tabular}%